\pdfoutput=1

\documentclass[11pt]{article}

\usepackage[]{ACL2023}

\usepackage{times}
\usepackage{latexsym}

\usepackage[T1]{fontenc}

\usepackage[utf8]{inputenc}

\usepackage{microtype}

\usepackage{inconsolata}

\usepackage{amsthm}
\usepackage{amssymb}
\usepackage{tipa}
\usepackage{ragged2e}

\usepackage{hyperref}       
\usepackage{url}            
\usepackage{booktabs}       
\usepackage{amsfonts}       
\usepackage{amsmath}
\usepackage{nicefrac}       
\usepackage{microtype}      
\usepackage{lipsum}
\usepackage{graphicx}
\graphicspath{{figure/}}     
\usepackage{caption}
\usepackage{subcaption}
\usepackage{multirow}
\usepackage{booktabs}

\usepackage[final]{pdfpages}
\usepackage{tabularx}
\usepackage{tkz-tab}
\usepackage{framed,multicol}
\usepackage[framemethod=tikz]{mdframed}
\usepackage{xcolor}

\usepackage{wasysym}

\usepackage{pgfplots}
\usepackage{tikz,pgfplots,pgfplotstable}

\usepackage{framed,multicol}
\usepackage[framemethod=tikz]{mdframed}

\usepgfplotslibrary{colorbrewer}
\pgfplotsset{compat = 1.15, cycle list/Set1-8} 
\usetikzlibrary{pgfplots.statistics, pgfplots.colorbrewer} 
\usepackage{pgfplotstable}
\usepackage{filecontents}

\definecolor{markercolor}{RGB}{124.9, 255, 160.65}
\pgfplotsset{compat=1.3}
\usetikzlibrary{positioning, fit, calc}  
\tikzset{block/.style={draw, thick, text width=2cm ,minimum height=1.3cm, align=center},   
	line/.style={-latex}     
} 

\tikzset{blocktext/.style={draw, thick, text width=5.2cm ,minimum height=1.3cm, align=center},   
	line/.style={-latex}     
} 

\title{Performance Comparison of Large Language Models on VNHSGE English Dataset: OpenAI ChatGPT, Microsoft Bing Chat, and Google Bard}

\author{Xuan-Quy DAO \\
  School of Engineering \\
  Eastern International University \\
  Binh Duong, Vietnam \\
  \texttt{quy.dao@eiu.edu.vn} \\
}

\begin{document}
\maketitle
\begin{abstract}

This paper presents a performance comparison of three large language models (LLMs), namely OpenAI ChatGPT, Microsoft Bing Chat (BingChat), and Google Bard, on the VNHSGE English dataset. The performance of BingChat, Bard, and ChatGPT (GPT-3.5) is 92.4\%, 86\%, and 79.2\%, respectively. The results show that BingChat is better than ChatGPT and Bard. Therefore, BingChat and Bard can replace ChatGPT while ChatGPT is not yet officially available in Vietnam. The results also indicate that BingChat, Bard and ChatGPT outperform Vietnamese students in English language proficiency. The findings of this study contribute to the understanding of the potential of LLMs in English language education. The remarkable performance of ChatGPT, BingChat, and Bard demonstrates their potential as effective tools for teaching and learning English at the high school level.

\end{abstract}

\section{Introduction}

Artificial Intelligence (AI) has revolutionized the field of education, by transforming the ways of learning and teaching. One of the most prominent advancements in this domain is the development of large language models (LLMs), such as ChatGPT\footnote{\href{https://chat.openai.com/}{https://chat.openai.com}}, BingChat\footnote{\href{https://www.bing.com/chat }{https://www.bing.com/chat}}, and Bard\footnote{\href{https://bard.google.com}{https://bard.google.com}}(developed by OpenAI, Microsoft, and Google, respectively). The development and continuous improvement of these LLMs have paved the way for AI to be integrated into various educational applications and domains~\cite{OpenAI_gpt_4_report}. These models have advanced conversational abilities, closely resembling human-like interactions. This capability holds great promise for educational settings, including the utilization of virtual assistants, chatbots, and online learning support systems~\cite{kasneci2023chatgpt}. 

Although these models offer tremendous potential for personalized and interactive learning experiences, creating new opportunities for educational innovation and transforming the way we teach and learn, the development and deployment of AI in education require careful consideration of the effectiveness of these AI-powered educational tools in different educational and social contexts~\cite{kung2023performance},~\cite{thorp2023chatgpt}.

Similar to other parts of the world, LLMs hold great potential in the field of education in Vietnam. However, it is essential to conduct thorough research and evaluation to comprehensively understand their capabilities and limitations in the specific context of Vietnamese education. ChatGPT scored well on the Vietnamese High School Graduation Examination (VNHSGE)~\cite{dao2023exa} and obtained a good score on an English test case~\cite{dao2023eng}. However, it is worth noting that ChatGPT has not been officially deployed in Vietnam. Meanwhile, Vietnamese students have access to BingChat and Bard, two other LLMs. Thus, it is crucial to evaluate the capabilities of BingChat and Bard within the context of Vietnamese education.

We focus on evaluating the performance of BingChat and Bard on the VNHSGE English dataset, and comparing them to ChatGPT. The objective of this evaluation is to determine whether BingChat and Bard can serve as potential alternatives to ChatGPT in English education at the high school level.  Thus, we ask the following research questions: 
\begin{itemize}
	\item Research Question 1 (RS1): What is the performance of ChatGPT, BingChat, and Bard on the VNHSGE English dataset at the high school level in Vietnam?
	\item Research Question 2 (RS2): How do the LLMs perform in comparison to Vietnamese students in English language proficiency at the high school level in Vietnam?
	\item Research Question 3 (RS3): What potential do LLMs hold for English language teaching and learning at the high school level in the context of Vietnam? 
\end{itemize}

The main contribution of this paper provides a comprehensive evaluation of the performance of LLMs, comparing them to human performance on the same tasks, and identifying the potential applications of LLMs in the context of English language education at the high school level in Vietnam. These findings can inform educators, researchers, and policymakers in making informed decisions regarding the integration of LLMs into the English language curriculum and instructional practices. 


\section{Related works}



\subsection{Large Language Models}

AI is now capable of understanding and engaging in human-like communication, thanks to recent advancements in LLMs. These breakthroughs have opened up new opportunities for their application in the field of education. LLMs have demonstrated immense potential in areas such as education, content development, and language translation. The two primary architectures of LLMs are BERT (Bidirectional Encoder Representations from Transformers) and GPT (Generative Pre-trained Transformer). In 2018, Google introduced BERT~\cite{devlin2018bert}, which has excelled in various natural language processing (NLP) tasks. Developed by OpenAI~\cite{alec2018improving}, the GPT algorithm was trained on extensive unlabeled text datasets. Building on Google's research, Facebook's RoBERTa~\cite{liu2019roberta} was introduced, and in 2019, Google released T5~\cite{raffel2020exploring}. The year 2020 witnessed the creation of GPT-3 by OpenAI~\cite{brown2020language}, showcasing exceptional performance in a wide range of NLP tasks. Recently, OpenAI developed GPT-4~\cite{OpenAI_gpt_4_report}, a text-to-text machine learning system capable of processing both text and image inputs. GPT-4 has demonstrated human-level performance in many professional and academic criteria, although its performance may not match that of humans in all contexts. These advancements highlight the progress made in LLMs and their potential impact in various domains.

\subsection{Evaluation of LLMs on English}

ChatGPT scored an average of 7.18 on the national high school exam in the Netherlands, which is equivalent to the average score of all students participating in the exams in the Netherlands~\cite{de2023can}. In other research, ChatGPT outperformed Vietnamese students by scoring 7.92 in the national high school graduation exam in Vietnam~\cite{dao2023eng}. Additionally, ChatGPT/GPT-4 has been shown to be as good as commercial translation products at translation~\cite{jiao2023chatgpt}, as well as having the ability to perform multilingual tasks~\cite{bang2023multitask}.


\section{Methods}

\subsection{Dataset}

The VNHSGE English dataset~\cite{dao2023vnhsge} consists of 2500 multiple-choice questions and comprises a diverse range of exam questions extracted from high school examinations, covering various topics and assessing different linguistic abilities. The dataset includes questions related to pronunciation and stress, vocabulary and grammar, communication skills, reading fill-in-the-blank, and reading comprehension. The evaluation dataset comprises 250 multiple-choice questions corresponding to 5 exams from the academic years 2019, 2020, 2021, 2022, and 2023.

\subsection{Prompt}

In this study, zero-shot learning was employed, where LLMs were tasked with answering questions directly without any prior data or example questions. For the VNHSGE English dataset D, let \{Q, S\} denote pairs where Q is the question and S is the ground truth solution. Additionally, let P denote the context of words. The LLM's answer A is determined by
\begin{equation}
	A = f(P, Q)
\end{equation}
where f is the LLM, which takes into account the context P and the question Q. The context P in this case is a specific structure that guides the LLM’s response. It instructs the LLM to provide the answer in the following format: \{ Choice: “A” or “B” or “C” or “D”;
	Explanation: Explain the answer;
	The question is: [the actual question] \}. By following this structure, the LLM generates its answer A, which can be evaluated and compared to the ground truth solution S.

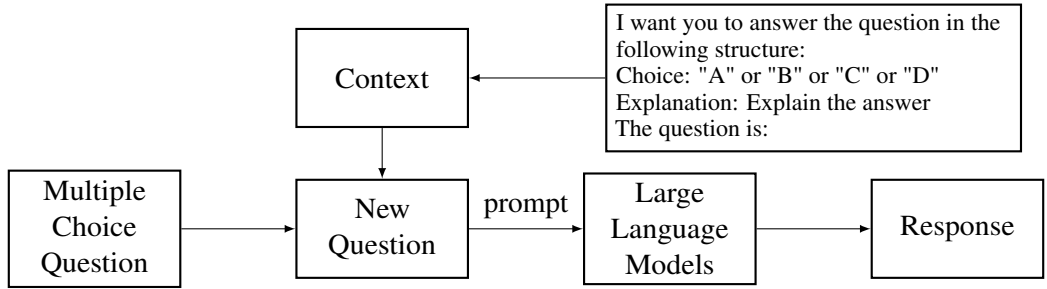
\begin{figure*}[ht!]
	\begin{center}
		\begin{tikzpicture}  
			[
			node distance = 1.5cm, 
			]
			\node[block] (a) {Multiple Choice Question};  
			\node[block,right=of a] (b) {New Question};   
			\node[block,right=of b] (c) {Large Language Models};  
			\node[block,right=of c] (d) {Response}; 
			\node[blocktext,align=flush left,font=\small] (f) at ([yshift=2cm]$(c)!0.5!(d)$) {I want you to answer the question in the following structure:  \\ 
				Choice: "A" or "B" or "C" or "D"  \\
				Explanation: Explain the answer \\
				The question is: {}
			};   
			\node[block] (h) at ([yshift=2cm]$(b)!1.0!(b)$) {Context}; 
			\draw[line] (a)-- (b);  
			\draw[line] (b)-- (c) node [midway, above, sloped] (TextNode) {prompt};  
			\draw[line] (c)-- (d);  
			\draw[line] (h)-- (b);  
			\draw[line] (f)-- (h);  
		\end{tikzpicture} 
	\end{center}
	\caption{Formatted question and LLMs response.}
	\label{fig:chatbot_response}
\end{figure*}

Figure~\ref{fig:chatbot_response} illustrates the process of prompting LLMs and retrieving the results. In the case of multiple-choice questions from the VNHSGE English dataset, the questions are formatted to align with the expected answer format. The questions are then sent to the LLMs' API.


\subsection{Grading}

To evaluate the performance of LLMs in answering questions, we assessed the LLM's response by comparing it to the ground truth. Given the question Q, the corresponding ground truth solution S, and the LLM's answer A, we assessed the LLM's response by comparing it to the ground truth solution S: 
\begin{equation}
	G = g(Q, S, A)  
\end{equation}
The evaluation process resulted in a binary grading system, where the LLM's answer was classified as correct or incorrect.

The answers of LLMs may be different because they are trained on different datasets. To capture the best and worst cases among the answers of LLMs, we introduced two variables: $\mathrm{LLM_{B}}$ represents the case with the highest graded answer among the LLMs, while $\mathrm{LLM_{W}}$ represents the case with the lowest graded answer among the LLMs
\begin{equation}
	\left\{\begin{matrix}
		\mathrm{LLM_{B}} = \max(G_i) \\
		\mathrm{LLM_{W}} = \min(G_i)
	\end{matrix}\right. 
\end{equation}
where $G_i$ represents the grading corresponding to ChatGPT, BingChat, and Bard. These variables allow us to analyze the upper and lower bounds of the performance of LLMs in terms of grading their answers.

\section{Results}

\subsection{Answers of LLMs}

In this section, we present the answers provided by ChatGPT, BingChat, and Bard on the VNHSGE English dataset. We provide illustrative examples for different types of questions, including pronunciation and stress, grammar and vocabulary, communication, reading fill-in-the-blank, and reading comprehension. For detailed information about the answers provided by the LLMs, please refer to Section \ref{sec:llm_a_appendix}.

\subsection{Performance of LLMs}

To answer RS~1, we evaluate the performance of ChatGPT, BingChat, and Bard on the VNHSGE English dataset. The findings provide insights into how accurate and effective these LLMs are in answering questions from the dataset. The contribution of this analysis is a comparative assessment of the performance of the three models, which can guide educators and researchers in selecting the most suitable LLM for English language tasks in the context of high school education.

\subsubsection{Question order}

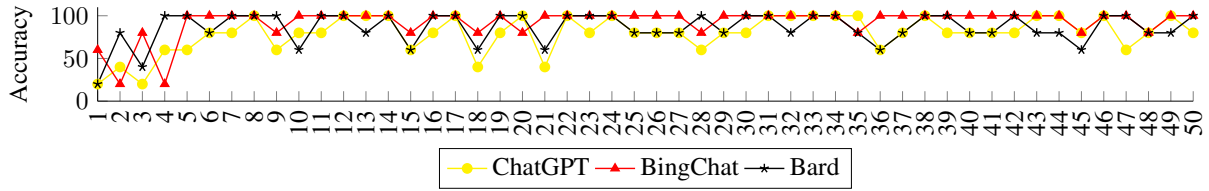
\begin{figure*}[h!]
	\begin{center}
		\resizebox{\textwidth}{!}{%
		\begin{tikzpicture}
			\begin{axis}
				[
				ylabel={Accuracy},
				legend style={at={(0.5,-0.5)}, 	
					anchor=north,legend columns=-1}, 
				symbolic x coords={
					1,
					2,
					3,
					4,
					5,
					6,
					7,
					8,
					9,
					10,
					11,
					12,
					13,
					14,
					15,
					16,
					17,
					18,
					19,
					20,
					21,
					22,
					23,
					24,
					25,
					26,
					27,
					28,
					29,
					30,
					31,
					32,
					33,
					34,
					35,
					36,
					37,
					38,
					39,
					40,
					41,
					42,
					43,
					44,
					45,
					46,
					47,
					48,
					49,
					50,
				},
				xtick=data,
				x tick label style={rotate=90,anchor=east},
				ymin=0,
				xmin=1,
				xmax=50,
				width=\textwidth, 
				height=3cm, 
				width=18cm,
				axis x line*=bottom, axis y line*=left
				]
				\addplot[color=yellow,mark=*,semithick, mark options={solid,}]
				coordinates{
					(1,20)
					(2,40)
					(3,20)
					(4,60)
					(5,60)
					(6,80)
					(7,80)
					(8,100)
					(9,60)
					(10,80)
					(11,80)
					(12,100)
					(13,100)
					(14,100)
					(15,60)
					(16,80)
					(17,100)
					(18,40)
					(19,80)
					(20,100)
					(21,40)
					(22,100)
					(23,80)
					(24,100)
					(25,80)
					(26,80)
					(27,80)
					(28,60)
					(29,80)
					(30,80)
					(31,100)
					(32,100)
					(33,100)
					(34,100)
					(35,100)
					(36,60)
					(37,80)
					(38,100)
					(39,80)
					(40,80)
					(41,80)
					(42,80)
					(43,100)
					(44,100)
					(45,80)
					(46,100)
					(47,60)
					(48,80)
					(49,100)
					(50,80)
				};
			
			\addplot[color=red,mark=triangle*,semithick, mark options={solid,}]
			coordinates{
				(1,60)
				(2,20)
				(3,80)
				(4,20)
				(5,100)
				(6,100)
				(7,100)
				(8,100)
				(9,80)
				(10,100)
				(11,100)
				(12,100)
				(13,100)
				(14,100)
				(15,80)
				(16,100)
				(17,100)
				(18,80)
				(19,100)
				(20,80)
				(21,100)
				(22,100)
				(23,100)
				(24,100)
				(25,100)
				(26,100)
				(27,100)
				(28,80)
				(29,100)
				(30,100)
				(31,100)
				(32,100)
				(33,100)
				(34,100)
				(35,80)
				(36,100)
				(37,100)
				(38,100)
				(39,100)
				(40,100)
				(41,100)
				(42,100)
				(43,100)
				(44,100)
				(45,80)
				(46,100)
				(47,100)
				(48,80)
				(49,100)
				(50,100)
				
			};
			
			\addplot[color=black,mark=star,semithick, mark options={solid,}]
			coordinates{
				(1,20)
				(2,80)
				(3,40)
				(4,100)
				(5,100)
				(6,80)
				(7,100)
				(8,100)
				(9,100)
				(10,60)
				(11,100)
				(12,100)
				(13,80)
				(14,100)
				(15,60)
				(16,100)
				(17,100)
				(18,60)
				(19,100)
				(20,100)
				(21,60)
				(22,100)
				(23,100)
				(24,100)
				(25,80)
				(26,80)
				(27,80)
				(28,100)
				(29,80)
				(30,100)
				(31,100)
				(32,80)
				(33,100)
				(34,100)
				(35,80)
				(36,60)
				(37,80)
				(38,100)
				(39,100)
				(40,80)
				(41,80)
				(42,100)
				(43,80)
				(44,80)
				(45,60)
				(46,100)
				(47,100)
				(48,80)
				(49,80)
				(50,100)
				
			};
			
			\legend{ChatGPT, BingChat, Bard}	
			\end{axis}
		\end{tikzpicture}
		}
	\end{center}
	\caption{Correctness average of ChatGPT, BingChat, and Bard in question order.}
	\label{fig:question_order}
\end{figure*}

Figure~\ref{fig:question_order} illustrates the accuracy of the answers of LLMs according to the order of the questions. The results show that the LLMs models do not provide accurate answers for the first four questions, which are related to phonetics. This may be because models like ChatGPT, BingChat, and Bard were not sufficiently trained in the field of phonetics. LLMs perform well on the remaining portions of the exam, proving that LLMs have received adequate training in English vocabulary, grammar, and reading comprehension.

\subsubsection{Performance evaluation}

Table~\ref{llm_performance} presents the performance of LLMs. BingChat demonstrates better results compared to ChatGPT and Bard. The accuracy of $\mathrm{LLM_{B}}$ is 97.2\%, demonstrating the significant potential of LLMs in English education at the high school level. 

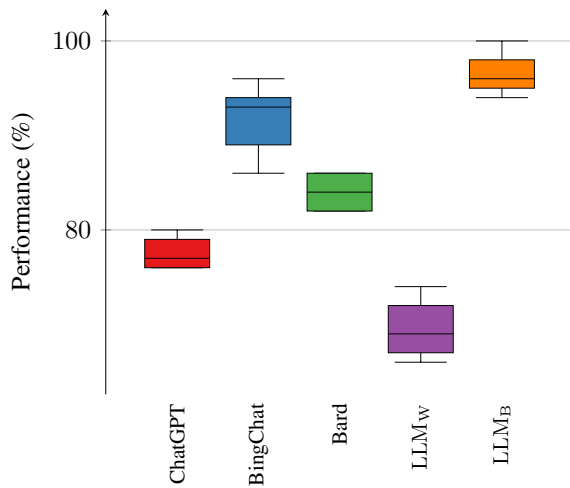
\begin{figure}[h!]
	\begin{center}
		\resizebox{0.475\textwidth}{!}{%
		\begin{tikzpicture}
			\pgfplotstableread[col sep=comma]{performance}\csvdata
			\pgfplotstabletranspose\datatransposed{\csvdata} 
			\begin{axis}[
				boxplot/draw direction = y,
				x axis line style = {opacity=0},
				axis x line* = bottom,
				axis y line = left,
				enlarge y limits,
				ymajorgrids,
				xtick = {1, 2, 3, 4, 5},
				xticklabel style = {align=center, font=\small, rotate=90},
				xticklabels = {ChatGPT, BingChat, Bard, $\mathrm{LLM_{W}}$, $\mathrm{LLM_{B}}$},
				xtick style = {draw=none}, 
				ylabel = {Performance (\%)},
				ytick = {0, 40, 60, 80, 100}
				]
				\foreach \n in {1,...,5} {
					\addplot+[boxplot, fill, draw=black] table[y index=\n] {\datatransposed};
				}
			\end{axis}
		\end{tikzpicture}
		}
		\caption{Stability of LLMs performance.}
		\label{fig:stability}
	\end{center}
\end{figure}

Figure~\ref{fig:stability} illustrates the stability of LLMs across the years 2019-2023. The performance indices $\mathrm{LLM_{B}}$, ChatGPT, and Bard demonstrate relatively stable results, while BingChat and $\mathrm{LLM_{W}}$ exhibit variations between academic years. The consistent performance of LLMs over the years indicates their ability to maintain stable performance. This demonstrates that LLMs are trained on a large amount of data and can be valuable tools in education. 

\begingroup
\renewcommand{\arraystretch}{1.5} 
\begin{table}[h]
	\caption{Performance ($\%$)}
	\label{llm_performance}
	\resizebox{0.475\textwidth}{!}{%
	\begin{tabular}{c|c|c|c|c|c|c|}
		\cline{2-7}
		& \textbf{2019} & \textbf{2020} & \textbf{2021} & \textbf{2022} & \textbf{2023}  & \textbf{AVG} \\ \hline
		\multicolumn{1}{|c|}{ChatGPT} & 76            & 86            & 76            & 80            & 78  & 79.2          \\ \hline
		\multicolumn{1}{|c|}{BingChat} & 92            & 96            & 86            & 94            & 94   & 92.4        \\ \hline
		\multicolumn{1}{|c|}{Bard} & 82            & 94            & 82            & 86            & 86        & 86    \\ \hline
		\multicolumn{1}{|c|}{$\mathrm{LLM_{W}}$} & 66            & 82            & 68            & 74           & 70 & 72           \\ \hline
		\multicolumn{1}{|c|}{$\mathrm{LLM_{B}}$} & 96            & 100            & 94            & 96            & 100  & 97.2          \\ \hline
	\end{tabular}
	}
\end{table}
\endgroup


\subsection{LLMs and Vietnamese students}

\subsubsection{LLMs perform better in English than Vietnamese students}

To answer RS~2, we examine the performance of the LLMs in comparison to Vietnamese students' English language skills. Our aim is to determine whether LLMs possess abilities comparable to those of humans, although this comparison is challenging due to the dissimilar settings. By conducting this comparison, we can evaluate whether LLMs can serve as effective study aids for Vietnamese students in various subject areas. 

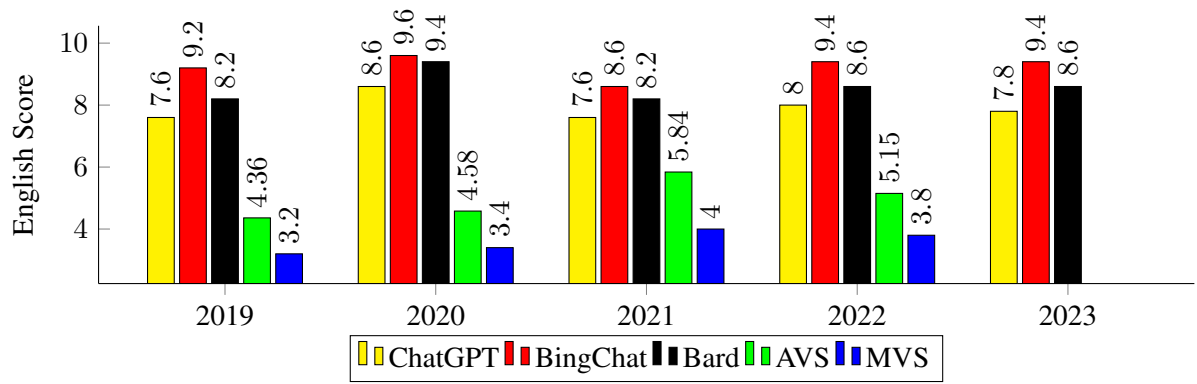
\begin{figure*}[ht!]
	\begin{center}
		\begin{tikzpicture}  
			\begin{axis}  
				[  
				ybar, 
				bar width=10pt,
				enlargelimits=0.15, 
				legend style={at={(0.5,-0.2)}, 	
					anchor=north,legend columns=-1},       
				ylabel={English Score}, 
				symbolic x coords={
					2019,
					2020,
					2021,
					2022,
					2023,
				},  
				xtick=data,  
				nodes near coords,  
				every node near coord/.append style={rotate=90, anchor=west},
				width=\textwidth, 
				height=5cm, 
				width=16cm,
				axis x line*=bottom, axis y line*=left
				]  
				\addplot [fill=yellow] coordinates {
					(2019,7.6)
					(2020,8.6)
					(2021,7.6)
					(2022,8)
					(2023,7.8)
				};
				\addplot [fill=red] coordinates {
					(2019,9.2)
					(2020,9.6)
					(2021,8.6)
					(2022,9.4)
					(2023,9.4)						
				};
				\addplot [fill=black] coordinates {
					(2019,8.2)
					(2020,9.4)
					(2021,8.2)
					(2022,8.6)
					(2023,8.6)						
				};
				\addplot [fill=green] coordinates {
					(2019,4.36)
					(2020,4.58)
					(2021,5.84)
					(2022,5.15)
					(2023,)							
				};
				\addplot [fill=blue] coordinates {
					(2019,3.2)
					(2020,3.4)
					(2021,4)
					(2022,3.8)
					(2023,)
				};
				\legend{ChatGPT, BingChat, Bard, AVS, MVS} 
			\end{axis}  
		\end{tikzpicture}  
	\end{center}
	\caption{Comparison of the performance of LLMs and Vietnamese students.}
	\label{fig:comparison}
\end{figure*}

The score distribution of the students' exams serves as an indicator for evaluating their performance. The Vietnam Ministry of Education and Training releases score distribution charts for each subject every year, which help assess the candidates' competence and determine the difficulty level of the exams. By gathering the score distributions (\href{https://moet.gov.vn/tintuc/Pages/tin-tong-hop.aspx?ItemID=6111}{2019},~
\href{https://moet.gov.vn/tintuc/Pages/tin-tong-hop.aspx?ItemID=6879}{2020},~
\href{https://moet.gov.vn/tintuc/Pages/tin-tong-hop.aspx?ItemID=7451}{2021},~
\href{https://vietnamnet.vn/pho-diem-cac-mon-thi-tot-nghiep-thpt-2022-2042421.html}{2022}), we can compare the performance of LLMs with that of Vietnamese students. To facilitate this comparison, we compare the score of LLMs to the average score (AVS) and the most reached score (MVS) by Vietnamese students. For instance, in 2023, the AVS and MVS were 5.84 and 3.8, respectively (see Appendix section~\ref{sec:spectrum_appendix} for comparison in Vietnamese score spectrum). Table~\ref{llm_score} shows the score of LLMs on a 10-point scale. Figure~\ref{fig:comparison} shows the effectiveness of LLMs with Vietnamese students.

\begingroup
\renewcommand{\arraystretch}{1.5} 
\begin{table}[h]
	\caption{Scores}
	\label{llm_score}
	\resizebox{0.475\textwidth}{!}{%
		\begin{tabular}{c|c|c|c|c|c|c|}
			\cline{2-7}
			& \textbf{2019} & \textbf{2020} & \textbf{2021} & \textbf{2022} & \textbf{2023} & \textbf{AVG} \\ \hline
			\multicolumn{1}{|c|}{ChatGPT} & 7.6            & 8.6            & 7.6            & 8.0            & 7.8   & 7.92         \\ \hline
			\multicolumn{1}{|c|}{BingChat} & 9.2            & 9.6            & 8.6            & 9.4            & 9.4   & 9.24         \\ \hline
			\multicolumn{1}{|c|}{Bard} & 8.2            & 9.4            & 8.2            & 8.6            & 8.6    & 8.6        \\ \hline
		\end{tabular}
	}
\end{table}
\endgroup

The results indicate that LLMs exhibit a higher level of English proficiency compared to Vietnamese students. One key advantage of LLMs is their ability to access vast amounts of information and data, including extensive language resources and pre-training on large corpora. This enables them to have a broader knowledge base and a deeper understanding of the English language. Consequently, LLMs demonstrate superior performance in answering questions and providing accurate responses compared to Vietnamese students.

\subsubsection{Potential of LLMs in helping Vietnamese students learn English}

To answer RS~3, we explore the potential of LLMs for English language teaching and learning at the high school level in Vietnam. By evaluating the performance of the LLMs on the VNHSGE English dataset, the research highlights the capabilities of these models in assisting with language instruction and learning. The findings contribute to identifying the specific areas where LLMs can be leveraged, such as vocabulary acquisition, grammar comprehension, and reading comprehension. The findings of the research can inform the development of educational strategies and materials that incorporate LLMs to enhance English language education in Vietnam. 

LLMs can indeed provide valuable assistance to Vietnamese students in learning English. Here are some ways LLMs can be helpful:

\begin{itemize}
	\item Language practice: LLMs can act as conversational partners, allowing Vietnamese students to practice their English communication skills. Students can engage in dialogue with the model, asking questions, discussing topics, and receiving responses that simulate real conversations. 
	
	\item Personalized learning: LLMs have the potential to adapt to individual student's needs and provide personalized learning experiences. By analyzing students' performance and identifying areas of improvement, LLMs can generate customized exercises, quizzes, or study materials tailored to each student's specific requirements.
	
	\item Language comprehension: LLMs can process and understand English text, including articles, books, and educational resources. Vietnamese students can use LLMs to improve their reading comprehension by getting instant definitions, explanations, and examples of unfamiliar words or phrases.
	
	\item Writing support: LLMs can assist students in improving their English writing skills. Students can use LLMs to generate suggestions for sentence structure, grammar, and vocabulary, and even receive feedback on their writing. This can help students enhance their written expression and produce more accurate and coherent English texts.
	
	\item Language resources: LLMs can serve as a vast repository of information, providing access to various English learning materials, such as grammar rules, idiomatic expressions, vocabulary lists, and sample essays. Students can leverage these resources to enhance their understanding of English language concepts and improve their overall language proficiency.
\end{itemize}


\section{Conclusion}

This study compared the performance of three LLMs (OpenAI ChatGPT, Microsoft Bing Chat, and Google Bard) on the VNHSGE English dataset. The results showed that these LLMs have the potential to be used in education, as they demonstrated high accuracy in answering multiple-choice questions and were consistent in their performance across different years of the dataset. 

Additionally, LLMs are significantly superior to Vietnamese students in English language proficiency suggesting that they have the potential to be valuable tools for facilitating English language learning among Vietnamese students. By leveraging the strengths of language models, educators, and students can benefit from personalized and adaptive learning experiences that meet individual needs and enhance overall language proficiency. This highlights the potential of language models to revolutionize English language education and empower Vietnamese students to excel in their language-learning endeavors.

\bibliography{paper_7_3}
\bibliographystyle{acl_natbib}

\appendix

\section{Answes of LLMs}
\label{sec:llm_a_appendix}

This section illustrates the answers of LLMs on the VNHSGE English dataset. 

\subsection{Pronunciation and stress question}

\begin{mdframed}[linewidth=1pt,linecolor=red] 
	\textbf{Question}: Mark the letter A, B, C, or D on your answer sheet to indicate the word whose underlined part differs from the other three in pronunciation in each of the following questions.
	\begin{choices}
		\choice maintain\underline{ed} 
		\choice promot\underline{ed} 
		\choice requir\underline{ed} 	
		\choice argu\underline{ed}
	\end{choices}
	
	\textbf{Solution}: 
	Pronounce the ending "ed"
	Explain:
	A. maintained /mein'teind/
	B. promoted /pr\textschwa'm\textschwa utid/
	C. required /rI'kwai\textschwa(r)d/
	D. say /'a:gju:d/
	How to pronounce the ending "ed":
	+ The ending "ed" is pronounced $/ \mathrm{t} /$ when the verb ends in $/ \mathrm{t} \int /, / \theta /, /$ $ \int /, / \mathrm{s} /, / \mathrm{k} /, / \mathrm{p} /, / \mathrm{f} / \mathrm{+}$ "ed" is pronounced $/ \mathrm{Id} /$ when the verb ends in $/ \mathrm{t} /, / \mathrm{d} /$
	+ The ending "ed" is pronounced/d/ for the rest of the cases Option $B$, the underlined part is pronounced /Id/, the rest is pronounced /d/.
	Choose \textcolor{red}{\textbf{B}}. 
	
	\textbf{ChatGPT}: Question 1: \textcolor{red}{\textbf{D}}. argu\underline{ed} 
	
	\textbf{BingChat}: For question 1, the word whose underlined part differs from the other three in pronunciation is \textcolor{red}{\textbf{D}}. argu\underline{ed}. 
	
	\textbf{Bard}: The answer is **\textcolor{red}{\textbf{A}}. maintain**.
	
	The other three words have the "ed" pronounced as /d/, but "maintain" has the "ed" pronounced as /t/.
		
\end{mdframed}

\subsection{Grammar and vocabulary questions}

\begin{mdframed}[linewidth=1pt,linecolor=red] 
	\textbf{Question}: Mark the letter A, B, C, or D on your answer sheet to indicate the correct answer to each of the following questions. 
	
	Question: He’s American,….?
	
	\begin{choices}
		\choice won’t he 
		\choice didn’t he 	
		\choice doesn’t he 
		\choice isn’t he
	\end{choices}
	
	\textbf{Solution}: Tag question. Explain:
	The verb in the main clause is "is", the subject is "He".
	=> Negative question tag “isn't he”.
	=> He's American, isn't he?
	Choose \textcolor{red}{\textbf{D}}. 
	
	\textbf{ChatGPT}: \textcolor{red}{\textbf{D}}. isn't he 
	
	\textbf{BingChat}: For question 5, the correct answer is \textcolor{red}{\textbf{D}}. isn’t he.
	
	\textbf{Bard}: The correct answer is \textcolor{red}{\textbf{D}}. The question is a tag question, which is a type of question that is used to confirm information. The tag question "isn't he?" is used to confirm that the person being referred to is American. The other options are not correct because they do not make sense in the context of the question.

\end{mdframed}

\begin{mdframed}[linewidth=1pt,linecolor=red] 
	\textbf{Question}: Mark the letter A, B, C, or D on your answer sheet to indicate the underlined part that needs correction in each of the following questions.
	
	Question: It is \underline{essential} to provide high school students with adequate career \underline{orientation}, so they can make more \underline{informative} decisions about their future \underline{major}.

	\begin{choices}
		\choice essential	
		\choice orientation	
		\choice informative	
		\choice major  
	\end{choices}
	
	\textbf{Solution}: Advanced Vocabulary.
	Explain:
	informative (adj): provides a lot of information, has the effect of improving knowledge
	informed (adj): to have a lot of knowledge or information about something
	Edit: “informative” => (informed)
	It is essential to provide high school students with adequate career guidance, so they can make more informed decisions about their future major. Choose \textcolor{red}{\textbf{C}}. 
	
	\textbf{ChatGPT}: It is essential to provide high school students with adequate career orientation, so they can make more informed decisions about their future major. No correction needed.
	
	\textbf{BingChat}: \textcolor{red}{\textbf{C}}. informative
	
	\textbf{Bard}: The correct answer is (\textcolor{red}{\textbf{D}}). The word major should be capitalized, as it is referring to a specific academic field of study. The other parts of the sentence are correct. 
	
\end{mdframed}

\subsection{Communication question}

\begin{mdframed}[linewidth=1pt,linecolor=red] 
	\textbf{Question}: Mark the letter A, B, C, or D on your answer sheet to indicate the option that best completes each of the following exchanges.
	
	Question 21: Jack and David are talking about taking a gap year.
	~  Jack: “I think taking a gap year is a waste of time.”
	~  David: “……. It gives gap-year takers a lot of valuable experiences.”

	\begin{choices}
		\choice I agree with you 	
		\choice I don’t quite agree   	
		\choice It’s right     		
		\choice My pleasure
	\end{choices}
	
	\textbf{Solution}: 
	Explanation:  A. I agree with you => not suitable meaning
	B. I don't quite agree
	C. That's right => doesn't match the meaning
	D. My pleasure => inappropriate meaning
	Choose \textcolor{red}{\textbf{B}}. 
	
	\textbf{ChatGPT}: Jack and David are talking about taking a gap year. ~ Jack: “I think taking a gap year is a waste of time.” ~ David: “I don’t quite agree. It gives gap-year takers a lot of valuable experiences.” \textcolor{red}{\textbf{B}}. “I don’t quite agree.
	
	\textbf{BingChat}: the option that best completes the exchange is \textcolor{red}{\textbf{B}}. I don’t quite agree. 
	
	\textbf{Bard}: 
	The correct answer is **\textcolor{red}{\textbf{B}}. I don’t quite agree**.
	David disagrees with Jack's opinion that taking a gap year is a waste of time. He believes that gap-year takers can gain a lot of valuable experiences, such as travel, work experience, and personal growth.

\end{mdframed}

\subsection{Reading Fill-in-the-Blank question}

\begin{mdframed}[linewidth=1pt,linecolor=red] 
	\textbf{Question}: Read the following passage and mark the letter A, B, C, or D on your answer sheet to indicate the correct word or phrase that bestfits each of the numbered blanks from 26 to 30.
	
	THE BEST STUDENTS
	
	According to most professors, mature students are ideal students because they are hard-working and become actively involved in all aspects of the learning process. The majority of mature students have a poor educational (26)…..but they manage to do exceptionally well at tertiary level. 
	 
	Question 26: 
	\begin{choices}
		\choice development 
		\choice background 
		\choice basement 		
		\choice institution
	\end{choices}
	
	\textbf{Solution}: The majority of mature students have a poor educational (26) background but they manage to do exceptionally  well at tertiary level. Choose \textcolor{red}{\textbf{B}}. 
	
	\textbf{ChatGPT}: \textcolor{red}{\textbf{B}}. background
	
	\textbf{BingChat}:  the correct word that best fits the blank is \textcolor{red}{\textbf{B}}. background. 
	
	\textbf{Bard}: The answer must be something that describes the educational history of mature students. Of the above choices, only background describes the educational history of mature students. So the answer is (\textcolor{red}{\textbf{B}}).
	Here are the other choices and why they are not the answer:
	* development: This does not describe the educational history of mature students.
	* basement: This is not a word that is typically used to describe educational history.
	* institution: This is a place where education takes place, but it does not describe the educational history of mature students.
		
\end{mdframed}

\subsection{Reading comprehension question}
\label{VNHSGE_Eng_RC}
\begin{mdframed}[linewidth=1pt,linecolor=red] 
	\textbf{Question}: 
	Read the following passage and mark the letter A, B, C, or D on your answer sheet to indicate the correct answer to each of the questions.
	
	Marcel Bich, a French manufacturer of traditional ink pens, was the man turning the ballpoint pen into an item that today almost anyone can afford. Bich was shocked at the poor quality of the ballpoint pens that were available, and also at their high cost. However, he recognised that the ballpoint was a firmly established invention, and he decided to design a cheap pen that worked well and would be commercially successful.
	
	Bich went to the Biro brothers and asked them if he could use the design of their original invention in one of his own pens. In return, he offered to pay them every time he sold a pen. Then, for two years, Bich studied the detailed construction of every ballpoint pen that was being sold, often working with a microscope.
	
	By 1950, he was ready to introduce his new wonder: a plastic pen with a clear barrel that wrote smoothly, did not leak and only cost a few cents. He called it the ‘Bic Cristal’. The ballpoint pen had finally become a \underline{practical} writing instrument. The public liked it immediately, and today it is as common as the pencil. In Britain, they are still called Biros, and many Bic models also say ‘Biro’ on the side of the pen, to remind people of their original inventors. 
	Bich became extremely wealthy thanks to his invention, \underline{which} had worldwide appeal. Over the next 60 years his company, Societe Bic, opened factories all over the world and expanded its range of inexpensive products. Today, Bic is as famous for its lighters and razors as it is for its pens, and you can even buy a Bic mobile phone.
		
	Question: The word “which” in paragraph 4 refers to…….
	\begin{choices}
		\choice factories 
		\choice company 
		\choice invention 
		\choice range
	\end{choices}
	
	\textbf{Solution}: 
	Explain:
	The word “which” in paragraph 4 refers to .....
	=> which replaces the noun that comes before it, "invention".
	Information: Bich became extremely wealthy thanks to his invention, which had worldwide appeal.. Choose \textcolor{red}{\textbf{C}}. 
	
	\textbf{ChatGPT}: \textcolor{red}{\textbf{C}}. invention.  
	
	\textbf{BingChat}:  the word "which" in paragraph 4 refers to \textcolor{red}{\textbf{C}}. invention .
	
	\textbf{Bard}: The word “which” in paragraph 4 refers to **\textcolor{red}{\textbf{C}}.invention**.
	
	In paragraph 4, it says:
	
	Bich became extremely wealthy thanks to his invention, which had worldwide appeal. Over the next 60 years his company, Societe Bic, opened factories all over the world and expanded its range of inexpensive products.
	
	The word “which” refers to the invention that Bich created. This invention was the ballpoint pen, which became a practical writing instrument that was affordable and widely used.
	.
	
\end{mdframed}

\section{Vietnamese score spectrum}
\label{sec:spectrum_appendix}

\begin{figure*}[h!]
	\begin{center}
		\begin{tikzpicture}
			\begin{axis}[
				legend style={at={(0.5,-0.125)}, 	
					anchor=north,legend columns=-1}, 
				symbolic x coords={
					0,
					0.2,
					0.4,
					0.6,
					0.8,
					1,
					1.2,
					1.4,
					1.6,
					1.8,
					2,
					2.2,
					2.4,
					2.6,
					2.8,
					3,
					3.2,
					3.4,
					3.6,
					3.8,
					4,
					4.2,
					4.4,
					4.6,
					4.8,
					5,
					5.2,
					5.4,
					5.6,
					5.8,
					6,
					6.2,
					6.4,
					6.6,
					6.8,
					7,
					7.2,
					7.4,
					7.6,
					7.8,
					8,
					8.2,
					8.4,
					8.6,
					8.8,
					9,
					9.2,
					9.4,
					9.6,
					9.8,
					10,
				},
				hide axis,
				ybar,
				bar width=5pt,
				ymin=0,
				ymax=10,
				every node near coord/.append style={rotate=90, anchor=west},
				width=\textwidth, 
				enlarge x limits={abs=0.5*\pgfplotbarwidth},
				height=9cm, 
				width=16cm,
				axis x line*=bottom, axis y line*=left
				]
				\addplot [fill=green] coordinates {
					(0,0)
				};
				\addplot [fill=red] coordinates {
					(5,0)
				};
				\addplot [fill=black] coordinates {
					(7,0)
				};	
				\addplot [fill=blue] coordinates {
					(10,0)
				};	
				\legend{ChatGPT, BingChat, Bard, Vietnamese students }	
			\end{axis}
			
			\begin{axis}[
				symbolic x coords={
					0,
					0.2,
					0.4,
					0.6,
					0.8,
					1,
					1.2,
					1.4,
					1.6,
					1.8,
					2,
					2.2,
					2.4,
					2.6,
					2.8,
					3,
					3.2,
					3.4,
					3.6,
					3.8,
					4,
					4.2,
					4.4,
					4.6,
					4.8,
					5,
					5.2,
					5.4,
					5.6,
					5.8,
					6,
					6.2,
					6.4,
					6.6,
					6.8,
					7,
					7.2,
					7.4,
					7.6,
					7.8,
					8,
					8.2,
					8.4,
					8.6,
					8.8,
					9,
					9.2,
					9.4,
					9.6,
					9.8,
					10,
				},
				hide axis,
				x tick label style={rotate=90,anchor=east},
				ybar,
				bar width=5pt,
				ymin=0,
				every node near coord/.append style={rotate=90, anchor=west},
				width=\textwidth, 
				enlarge x limits={abs=0.5*\pgfplotbarwidth},
				height=9cm, 
				width=16cm,
				axis x line*=bottom, axis y line*=left
				]
				\addplot [fill=green] coordinates {
					(0,0)
					(0.2,0)
					(0.4,0)
					(0.6,0)
					(0.8,0)
					(1,0)
					(1.2,0)
					(1.4,0)
					(1.6,0)
					(1.8,0)
					(2,0)
					(2.2,0)
					(2.4,0)
					(2.6,0)
					(2.8,0)
					(3,0)
					(3.2,0)
					(3.4,0)
					(3.6,0)
					(3.8,0)
					(4,0)
					(4.2,0)
					(4.4,0)
					(4.6,0)
					(4.8,0)
					(5,0)
					(5.2,0)
					(5.4,0)
					(5.6,0)
					(5.8,0)
					(6,0)
					(6.2,0)
					(6.4,0)
					(6.6,0)
					(6.8,0)
					(7,0)
					(7.2,0)
					(7.4,0)
					(7.6,55000)
					(7.8,0)
					(8,0)
					(8.2,0)
					(8.4,0)
					(8.6,0)
					(8.8,0)
					(9,0)
					(9.2,0)
					(9.4,0)
					(9.6,0)
					(9.8,0)
					(10,0)
					
				};	
			\end{axis}
			
			\begin{axis}[ 
				symbolic x coords={
					0,
					0.2,
					0.4,
					0.6,
					0.8,
					1,
					1.2,
					1.4,
					1.6,
					1.8,
					2,
					2.2,
					2.4,
					2.6,
					2.8,
					3,
					3.2,
					3.4,
					3.6,
					3.8,
					4,
					4.2,
					4.4,
					4.6,
					4.8,
					5,
					5.2,
					5.4,
					5.6,
					5.8,
					6,
					6.2,
					6.4,
					6.6,
					6.8,
					7,
					7.2,
					7.4,
					7.6,
					7.8,
					8,
					8.2,
					8.4,
					8.6,
					8.8,
					9,
					9.2,
					9.4,
					9.6,
					9.8,
					10,
				},
				hide axis,
				ybar,
				bar width=5pt,
				ymin=0,
				every node near coord/.append style={rotate=90, anchor=west},
				width=\textwidth, 
				enlarge x limits={abs=0.5*\pgfplotbarwidth},
				height=9cm, 
				width=16cm,
				axis x line*=bottom, axis y line*=left
				]
				\addplot [fill=red] coordinates {
					(0,0)
					(0.2,0)
					(0.4,0)
					(0.6,0)
					(0.8,0)
					(1,0)
					(1.2,0)
					(1.4,0)
					(1.6,0)
					(1.8,0)
					(2,0)
					(2.2,0)
					(2.4,0)
					(2.6,0)
					(2.8,0)
					(3,0)
					(3.2,0)
					(3.4,0)
					(3.6,0)
					(3.8,0)
					(4,0)
					(4.2,0)
					(4.4,0)
					(4.6,0)
					(4.8,0)
					(5,0)
					(5.2,0)
					(5.4,0)
					(5.6,0)
					(5.8,0)
					(6,0)
					(6.2,0)
					(6.4,0)
					(6.6,0)
					(6.8,0)
					(7,0)
					(7.2,0)
					(7.4,0)
					(7.6,0)
					(7.8,0)
					(8,0)
					(8.2,0)
					(8.4,0)
					(8.6,55000)
					(8.8,0)
					(9,0)
					(9.2,0)
					(9.4,0)
					(9.6,0)
					(9.8,0)
					(10,0)
				};	
			\end{axis}
			
			\begin{axis}[ 
				symbolic x coords={
					0,
					0.2,
					0.4,
					0.6,
					0.8,
					1,
					1.2,
					1.4,
					1.6,
					1.8,
					2,
					2.2,
					2.4,
					2.6,
					2.8,
					3,
					3.2,
					3.4,
					3.6,
					3.8,
					4,
					4.2,
					4.4,
					4.6,
					4.8,
					5,
					5.2,
					5.4,
					5.6,
					5.8,
					6,
					6.2,
					6.4,
					6.6,
					6.8,
					7,
					7.2,
					7.4,
					7.6,
					7.8,
					8,
					8.2,
					8.4,
					8.6,
					8.8,
					9,
					9.2,
					9.4,
					9.6,
					9.8,
					10,
				},
				hide axis,
				ybar,
				bar width=5pt,
				ymin=0,
				every node near coord/.append style={rotate=90, anchor=west},
				width=\textwidth, 
				enlarge x limits={abs=0.5*\pgfplotbarwidth},
				height=9cm, 
				width=16cm,
				axis x line*=bottom, axis y line*=left
				]
				\addplot [fill=black] coordinates {
					(0,0)
					(0.2,0)
					(0.4,0)
					(0.6,0)
					(0.8,0)
					(1,0)
					(1.2,0)
					(1.4,0)
					(1.6,0)
					(1.8,0)
					(2,0)
					(2.2,0)
					(2.4,0)
					(2.6,0)
					(2.8,0)
					(3,0)
					(3.2,0)
					(3.4,0)
					(3.6,0)
					(3.8,0)
					(4,0)
					(4.2,0)
					(4.4,0)
					(4.6,0)
					(4.8,0)
					(5,0)
					(5.2,0)
					(5.4,0)
					(5.6,0)
					(5.8,0)
					(6,0)
					(6.2,0)
					(6.4,0)
					(6.6,0)
					(6.8,0)
					(7,0)
					(7.2,0)
					(7.4,0)
					(7.6,0)
					(7.8,0)
					(8,0)
					(8.2,55000)
					(8.4,0)
					(8.6,0)
					(8.8,0)
					(9,0)
					(9.2,0)
					(9.4,0)
					(9.6,0)
					(9.8,0)
					(10,0)
				};	
			\end{axis}
		
			\begin{axis}[
				ylabel={Number of Student},
				symbolic x coords={
					0,
					0.2,
					0.4,
					0.6,
					0.8,
					1,
					1.2,
					1.4,
					1.6,
					1.8,
					2,
					2.2,
					2.4,
					2.6,
					2.8,
					3,
					3.2,
					3.4,
					3.6,
					3.8,
					4,
					4.2,
					4.4,
					4.6,
					4.8,
					5,
					5.2,
					5.4,
					5.6,
					5.8,
					6,
					6.2,
					6.4,
					6.6,
					6.8,
					7,
					7.2,
					7.4,
					7.6,
					7.8,
					8,
					8.2,
					8.4,
					8.6,
					8.8,
					9,
					9.2,
					9.4,
					9.6,
					9.8,
					10,
				},
				xtick=data,
				x tick label style={rotate=90,anchor=east},
				ybar,
				bar width=5pt,
				ymin=0,
				nodes near coords,   
				every node near coord/.append style={rotate=90, anchor=west},
				width=\textwidth, 
				enlarge x limits={abs=0.5*\pgfplotbarwidth},
				height=9cm, 
				width=16cm,
				axis x line*=bottom, axis y line*=left
				]
				\addplot [fill=blue] coordinates {
					(0,1)
					(0.2,0)
					(0.4,0)
					(0.6,3)
					(0.8,34)
					(1,106)
					(1.2,361)
					(1.4,713)
					(1.6,1619)
					(1.8,3302)
					(2,5868)
					(2.2,8732)
					(2.4,12280)
					(2.6,16517)
					(2.8,20491)
					(3,23980)
					(3.2,26527)
					(3.4,28537)
					(3.6,29183)
					(3.8,29498)
					(4,29504)
					(4.2,28943)
					(4.4,28317)
					(4.6,27791)
					(4.8,26867)
					(5,25860)
					(5.2,24631)
					(5.4,23337)
					(5.6,22660)
					(5.8,21964)
					(6,21090)
					(6.2,20102)
					(6.4,19403)
					(6.6,18911)
					(6.8,18665)
					(7,18354)
					(7.2,18464)
					(7.4,18219)
					(7.6,18498)
					(7.8,18915)
					(8,19319)
					(8.2,20258)
					(8.4,21176)
					(8.6,22490)
					(8.8,23724)
					(9,24471)
					(9.2,24251)
					(9.4,21582)
					(9.6,16586)
					(9.8,10543)
					(10,4345)
					
				};	
				
			\end{axis}
		\end{tikzpicture}
	\end{center}
	\caption{English score spectrum of Vietnamese students in 2021.}
	\label{fig:English_2021}
\end{figure*}
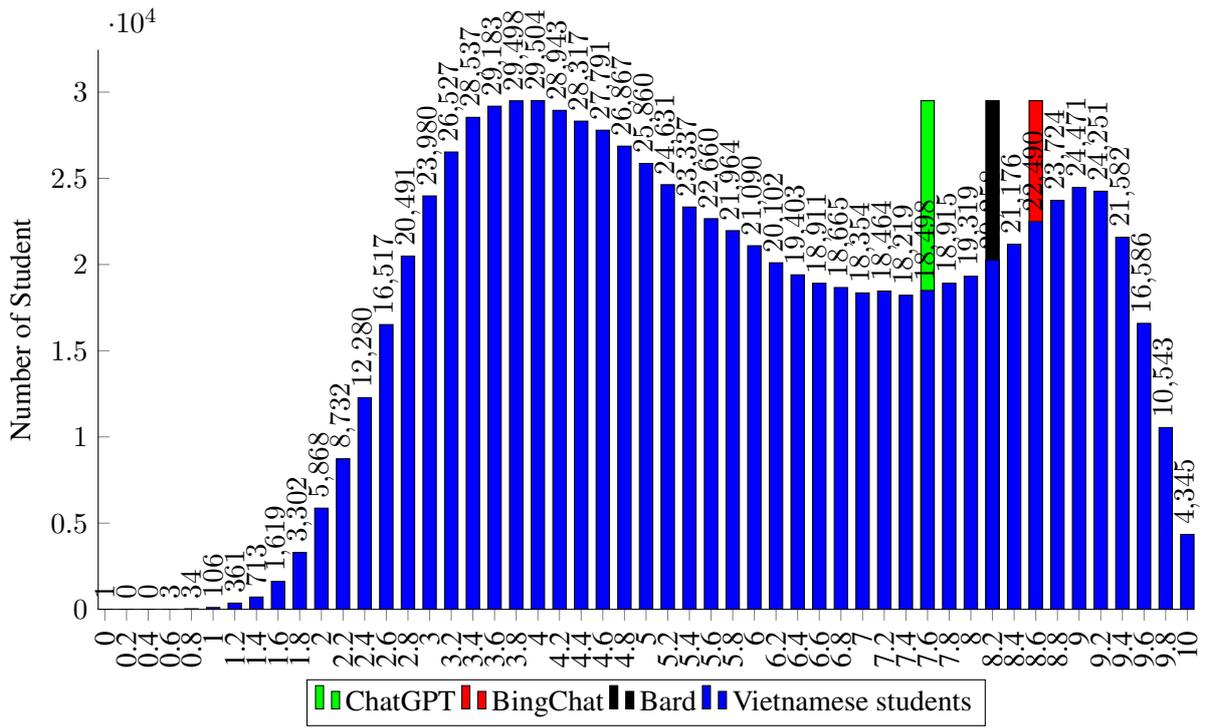

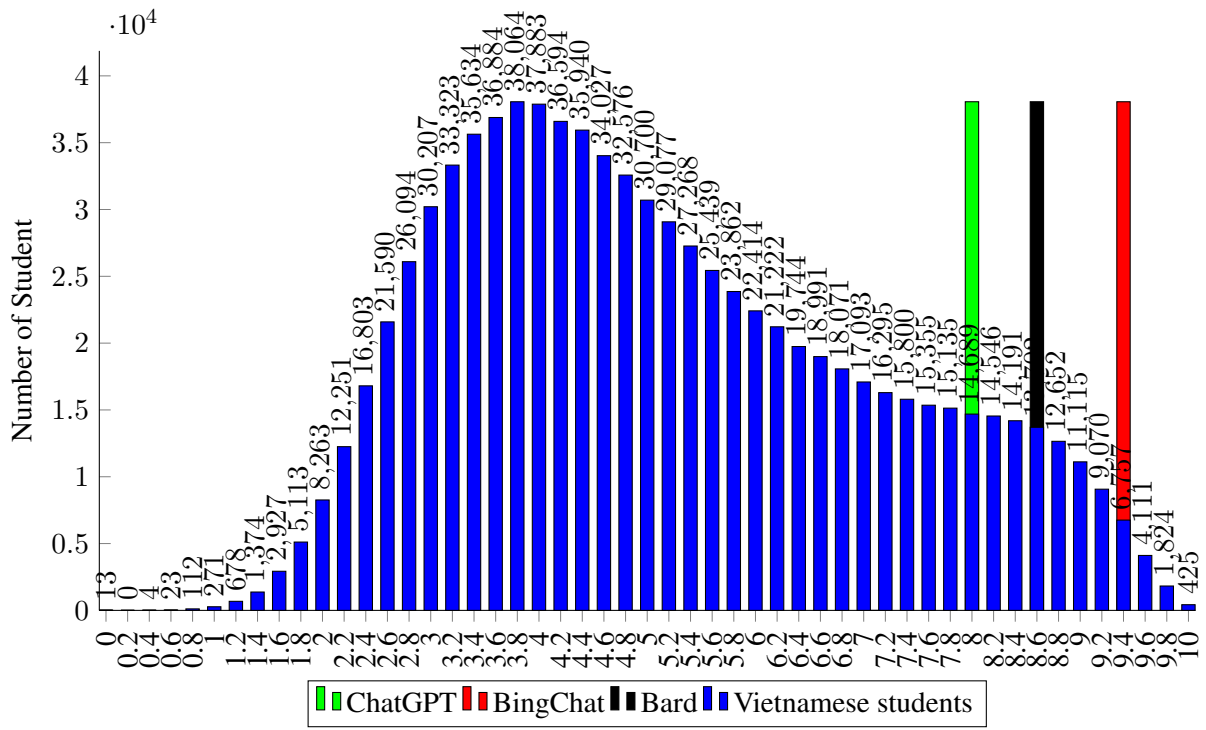
\begin{figure*}[h!]
	\begin{center}
		\begin{tikzpicture}
			\begin{axis}[
				legend style={at={(0.5,-0.125)}, 	
					anchor=north,legend columns=-1}, 
				symbolic x coords={
					0,
					0.2,
					0.4,
					0.6,
					0.8,
					1,
					1.2,
					1.4,
					1.6,
					1.8,
					2,
					2.2,
					2.4,
					2.6,
					2.8,
					3,
					3.2,
					3.4,
					3.6,
					3.8,
					4,
					4.2,
					4.4,
					4.6,
					4.8,
					5,
					5.2,
					5.4,
					5.6,
					5.8,
					6,
					6.2,
					6.4,
					6.6,
					6.8,
					7,
					7.2,
					7.4,
					7.6,
					7.8,
					8,
					8.2,
					8.4,
					8.6,
					8.8,
					9,
					9.2,
					9.4,
					9.6,
					9.8,
					10,
				},
				hide axis,
				ybar,
				bar width=5pt,
				ymin=0,
				ymax=10,
				every node near coord/.append style={rotate=90, anchor=west},
				width=\textwidth, 
				enlarge x limits={abs=0.5*\pgfplotbarwidth},
				height=9cm, 
				width=16cm,
				axis x line*=bottom, axis y line*=left
				]
				\addplot [fill=green] coordinates {
					(0,0)
				};
				\addplot [fill=red] coordinates {
					(5,0)
				};
				\addplot [fill=black] coordinates {
					(7,0)
				};	
				\addplot [fill=blue] coordinates {
					(10,0)
				};	
				\legend{ChatGPT, BingChat, Bard, Vietnamese students }	
			\end{axis}
			
			\begin{axis}[
				symbolic x coords={
					0,
					0.2,
					0.4,
					0.6,
					0.8,
					1,
					1.2,
					1.4,
					1.6,
					1.8,
					2,
					2.2,
					2.4,
					2.6,
					2.8,
					3,
					3.2,
					3.4,
					3.6,
					3.8,
					4,
					4.2,
					4.4,
					4.6,
					4.8,
					5,
					5.2,
					5.4,
					5.6,
					5.8,
					6,
					6.2,
					6.4,
					6.6,
					6.8,
					7,
					7.2,
					7.4,
					7.6,
					7.8,
					8,
					8.2,
					8.4,
					8.6,
					8.8,
					9,
					9.2,
					9.4,
					9.6,
					9.8,
					10,
				},
				hide axis,
				x tick label style={rotate=90,anchor=east},
				ybar,
				bar width=5pt,
				ymin=0,
				every node near coord/.append style={rotate=90, anchor=west},
				width=\textwidth, 
				enlarge x limits={abs=0.5*\pgfplotbarwidth},
				height=9cm, 
				width=16cm,
				axis x line*=bottom, axis y line*=left
				]
				\addplot [fill=green] coordinates {
					(0,0)
					(0.2,0)
					(0.4,0)
					(0.6,0)
					(0.8,0)
					(1,0)
					(1.2,0)
					(1.4,0)
					(1.6,0)
					(1.8,0)
					(2,0)
					(2.2,0)
					(2.4,0)
					(2.6,0)
					(2.8,0)
					(3,0)
					(3.2,0)
					(3.4,0)
					(3.6,0)
					(3.8,0)
					(4,0)
					(4.2,0)
					(4.4,0)
					(4.6,0)
					(4.8,0)
					(5,0)
					(5.2,0)
					(5.4,0)
					(5.6,0)
					(5.8,0)
					(6,0)
					(6.2,0)
					(6.4,0)
					(6.6,0)
					(6.8,0)
					(7,0)
					(7.2,0)
					(7.4,0)
					(7.6,0)
					(7.8,0)
					(8,55000)
					(8.2,0)
					(8.4,0)
					(8.6,0)
					(8.8,0)
					(9,0)
					(9.2,0)
					(9.4,0)
					(9.6,0)
					(9.8,0)
					(10,0)
					
				};	
			\end{axis}
			
			\begin{axis}[ 
				symbolic x coords={
					0,
					0.2,
					0.4,
					0.6,
					0.8,
					1,
					1.2,
					1.4,
					1.6,
					1.8,
					2,
					2.2,
					2.4,
					2.6,
					2.8,
					3,
					3.2,
					3.4,
					3.6,
					3.8,
					4,
					4.2,
					4.4,
					4.6,
					4.8,
					5,
					5.2,
					5.4,
					5.6,
					5.8,
					6,
					6.2,
					6.4,
					6.6,
					6.8,
					7,
					7.2,
					7.4,
					7.6,
					7.8,
					8,
					8.2,
					8.4,
					8.6,
					8.8,
					9,
					9.2,
					9.4,
					9.6,
					9.8,
					10,
				},
				hide axis,
				ybar,
				bar width=5pt,
				ymin=0,
				every node near coord/.append style={rotate=90, anchor=west},
				width=\textwidth, 
				enlarge x limits={abs=0.5*\pgfplotbarwidth},
				height=9cm, 
				width=16cm,
				axis x line*=bottom, axis y line*=left
				]
				\addplot [fill=red] coordinates {
					(0,0)
					(0.2,0)
					(0.4,0)
					(0.6,0)
					(0.8,0)
					(1,0)
					(1.2,0)
					(1.4,0)
					(1.6,0)
					(1.8,0)
					(2,0)
					(2.2,0)
					(2.4,0)
					(2.6,0)
					(2.8,0)
					(3,0)
					(3.2,0)
					(3.4,0)
					(3.6,0)
					(3.8,0)
					(4,0)
					(4.2,0)
					(4.4,0)
					(4.6,0)
					(4.8,0)
					(5,0)
					(5.2,0)
					(5.4,0)
					(5.6,0)
					(5.8,0)
					(6,0)
					(6.2,0)
					(6.4,0)
					(6.6,0)
					(6.8,0)
					(7,0)
					(7.2,0)
					(7.4,0)
					(7.6,0)
					(7.8,0)
					(8,0)
					(8.2,0)
					(8.4,0)
					(8.6,0)
					(8.8,0)
					(9,0)
					(9.2,0)
					(9.4,55000)
					(9.6,0)
					(9.8,0)
					(10,0)
				};	
			\end{axis}
			
				\begin{axis}[ 
				symbolic x coords={
					0,
					0.2,
					0.4,
					0.6,
					0.8,
					1,
					1.2,
					1.4,
					1.6,
					1.8,
					2,
					2.2,
					2.4,
					2.6,
					2.8,
					3,
					3.2,
					3.4,
					3.6,
					3.8,
					4,
					4.2,
					4.4,
					4.6,
					4.8,
					5,
					5.2,
					5.4,
					5.6,
					5.8,
					6,
					6.2,
					6.4,
					6.6,
					6.8,
					7,
					7.2,
					7.4,
					7.6,
					7.8,
					8,
					8.2,
					8.4,
					8.6,
					8.8,
					9,
					9.2,
					9.4,
					9.6,
					9.8,
					10,
				},
				hide axis,
				ybar,
				bar width=5pt,
				ymin=0,
				every node near coord/.append style={rotate=90, anchor=west},
				width=\textwidth, 
				enlarge x limits={abs=0.5*\pgfplotbarwidth},
				height=9cm, 
				width=16cm,
				axis x line*=bottom, axis y line*=left
				]
				\addplot [fill=black] coordinates {
					(0,0)
					(0.2,0)
					(0.4,0)
					(0.6,0)
					(0.8,0)
					(1,0)
					(1.2,0)
					(1.4,0)
					(1.6,0)
					(1.8,0)
					(2,0)
					(2.2,0)
					(2.4,0)
					(2.6,0)
					(2.8,0)
					(3,0)
					(3.2,0)
					(3.4,0)
					(3.6,0)
					(3.8,0)
					(4,0)
					(4.2,0)
					(4.4,0)
					(4.6,0)
					(4.8,0)
					(5,0)
					(5.2,0)
					(5.4,0)
					(5.6,0)
					(5.8,0)
					(6,0)
					(6.2,0)
					(6.4,0)
					(6.6,0)
					(6.8,0)
					(7,0)
					(7.2,0)
					(7.4,0)
					(7.6,0)
					(7.8,0)
					(8,0)
					(8.2,0)
					(8.4,0)
					(8.6,55000)
					(8.8,0)
					(9,0)
					(9.2,0)
					(9.4,0)
					(9.6,0)
					(9.8,0)
					(10,0)
				};	
			\end{axis}
		
			\begin{axis}[
				ylabel={Number of Student},
				symbolic x coords={
					0,
					0.2,
					0.4,
					0.6,
					0.8,
					1,
					1.2,
					1.4,
					1.6,
					1.8,
					2,
					2.2,
					2.4,
					2.6,
					2.8,
					3,
					3.2,
					3.4,
					3.6,
					3.8,
					4,
					4.2,
					4.4,
					4.6,
					4.8,
					5,
					5.2,
					5.4,
					5.6,
					5.8,
					6,
					6.2,
					6.4,
					6.6,
					6.8,
					7,
					7.2,
					7.4,
					7.6,
					7.8,
					8,
					8.2,
					8.4,
					8.6,
					8.8,
					9,
					9.2,
					9.4,
					9.6,
					9.8,
					10,
				},
				xtick=data,
				x tick label style={rotate=90,anchor=east},
				ybar,
				bar width=5pt,
				ymin=0,
				nodes near coords,   
				every node near coord/.append style={rotate=90, anchor=west},
				width=\textwidth, 
				enlarge x limits={abs=0.5*\pgfplotbarwidth},
				height=9cm, 
				width=16cm,
				axis x line*=bottom, axis y line*=left
				]
				\addplot [fill=blue] coordinates {
					(0,13)
					(0.2,0)
					(0.4,4)
					(0.6,23)
					(0.8,112)
					(1,271)
					(1.2,678)
					(1.4,1374)
					(1.6,2927)
					(1.8,5113)
					(2,8263)
					(2.2,12251)
					(2.4,16803)
					(2.6,21590)
					(2.8,26094)
					(3,30207)
					(3.2,33323)
					(3.4,35634)
					(3.6,36884)
					(3.8,38064)
					(4,37883)
					(4.2,36594)
					(4.4,35940)
					(4.6,34027)
					(4.8,32576)
					(5,30700)
					(5.2,29077)
					(5.4,27268)
					(5.6,25439)
					(5.8,23862)
					(6,22414)
					(6.2,21222)
					(6.4,19744)
					(6.6,18991)
					(6.8,18071)
					(7,17093)
					(7.2,16295)
					(7.4,15800)
					(7.6,15355)
					(7.8,15135)
					(8,14689)
					(8.2,14546)
					(8.4,14191)
					(8.6,13702)
					(8.8,12652)
					(9,11115)
					(9.2,9070)
					(9.4,6757)
					(9.6,4111)
					(9.8,1824)
					(10,425)			
				};	
				
			\end{axis}
		\end{tikzpicture}
	\end{center}
	\caption{English score spectrum of Vietnamese students in 2022.}
	\label{fig:English_2022}
\end{figure*}

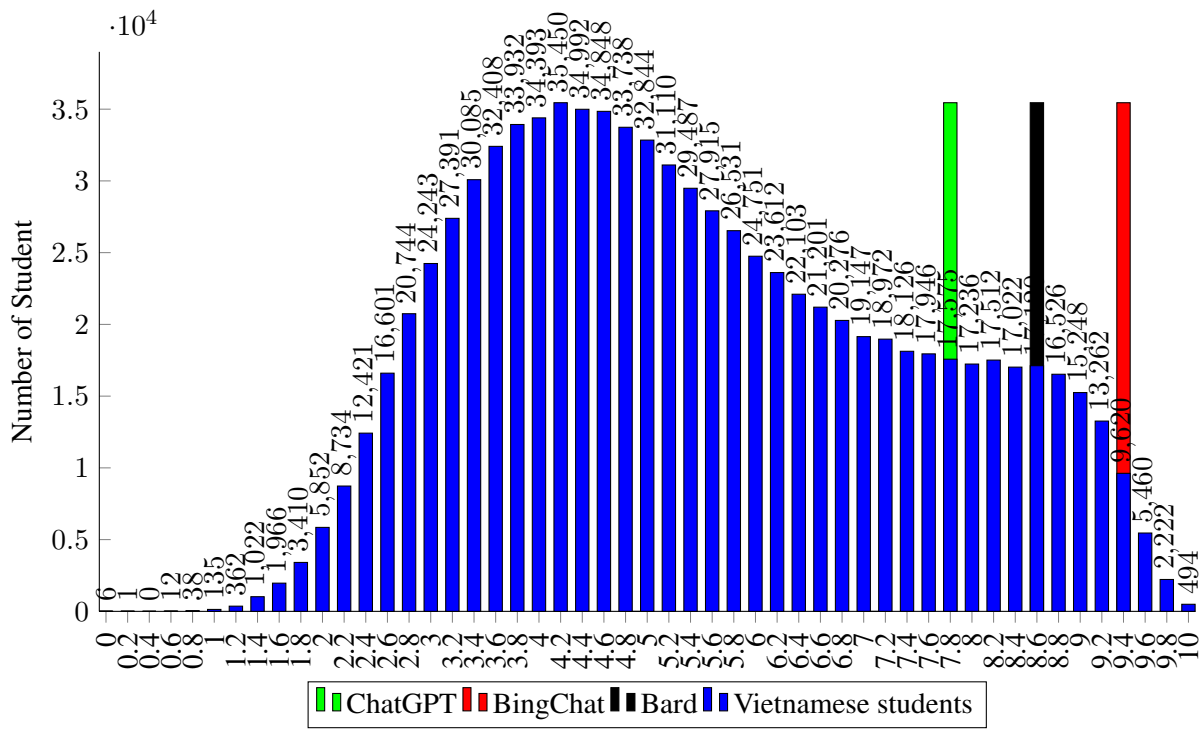
\begin{figure*}[h!]
	\begin{center}
		\begin{tikzpicture}
			\begin{axis}[
				legend style={at={(0.5,-0.125)}, 	
					anchor=north,legend columns=-1}, 
				symbolic x coords={
					0,
					0.2,
					0.4,
					0.6,
					0.8,
					1,
					1.2,
					1.4,
					1.6,
					1.8,
					2,
					2.2,
					2.4,
					2.6,
					2.8,
					3,
					3.2,
					3.4,
					3.6,
					3.8,
					4,
					4.2,
					4.4,
					4.6,
					4.8,
					5,
					5.2,
					5.4,
					5.6,
					5.8,
					6,
					6.2,
					6.4,
					6.6,
					6.8,
					7,
					7.2,
					7.4,
					7.6,
					7.8,
					8,
					8.2,
					8.4,
					8.6,
					8.8,
					9,
					9.2,
					9.4,
					9.6,
					9.8,
					10,
				},
				hide axis,
				ybar,
				bar width=5pt,
				ymin=0,
				ymax=10,
				every node near coord/.append style={rotate=90, anchor=west},
				width=\textwidth, 
				enlarge x limits={abs=0.5*\pgfplotbarwidth},
				height=9cm, 
				width=16cm,
				axis x line*=bottom, axis y line*=left
				]
				\addplot [fill=green] coordinates {
					(0,0)
				};
				\addplot [fill=red] coordinates {
					(5,0)
				};
				\addplot [fill=black] coordinates {
					(7,0)
				};	
				\addplot [fill=blue] coordinates {
					(10,0)
				};	
				\legend{ChatGPT, BingChat, Bard, Vietnamese students }	
			\end{axis}
			
			\begin{axis}[
				symbolic x coords={
					0,
					0.2,
					0.4,
					0.6,
					0.8,
					1,
					1.2,
					1.4,
					1.6,
					1.8,
					2,
					2.2,
					2.4,
					2.6,
					2.8,
					3,
					3.2,
					3.4,
					3.6,
					3.8,
					4,
					4.2,
					4.4,
					4.6,
					4.8,
					5,
					5.2,
					5.4,
					5.6,
					5.8,
					6,
					6.2,
					6.4,
					6.6,
					6.8,
					7,
					7.2,
					7.4,
					7.6,
					7.8,
					8,
					8.2,
					8.4,
					8.6,
					8.8,
					9,
					9.2,
					9.4,
					9.6,
					9.8,
					10,
				},
				hide axis,
				x tick label style={rotate=90,anchor=east},
				ybar,
				bar width=5pt,
				ymin=0,
				every node near coord/.append style={rotate=90, anchor=west},
				width=\textwidth, 
				enlarge x limits={abs=0.5*\pgfplotbarwidth},
				height=9cm, 
				width=16cm,
				axis x line*=bottom, axis y line*=left
				]
				\addplot [fill=green] coordinates {
					(0,0)
					(0.2,0)
					(0.4,0)
					(0.6,0)
					(0.8,0)
					(1,0)
					(1.2,0)
					(1.4,0)
					(1.6,0)
					(1.8,0)
					(2,0)
					(2.2,0)
					(2.4,0)
					(2.6,0)
					(2.8,0)
					(3,0)
					(3.2,0)
					(3.4,0)
					(3.6,0)
					(3.8,0)
					(4,0)
					(4.2,0)
					(4.4,0)
					(4.6,0)
					(4.8,0)
					(5,0)
					(5.2,0)
					(5.4,0)
					(5.6,0)
					(5.8,0)
					(6,0)
					(6.2,0)
					(6.4,0)
					(6.6,0)
					(6.8,0)
					(7,0)
					(7.2,0)
					(7.4,0)
					(7.6,0)
					(7.8,55000)
					(8,0)
					(8.2,0)
					(8.4,0)
					(8.6,0)
					(8.8,0)
					(9,0)
					(9.2,0)
					(9.4,0)
					(9.6,0)
					(9.8,0)
					(10,0)
					
				};	
			\end{axis}
			
			\begin{axis}[ 
				symbolic x coords={
					0,
					0.2,
					0.4,
					0.6,
					0.8,
					1,
					1.2,
					1.4,
					1.6,
					1.8,
					2,
					2.2,
					2.4,
					2.6,
					2.8,
					3,
					3.2,
					3.4,
					3.6,
					3.8,
					4,
					4.2,
					4.4,
					4.6,
					4.8,
					5,
					5.2,
					5.4,
					5.6,
					5.8,
					6,
					6.2,
					6.4,
					6.6,
					6.8,
					7,
					7.2,
					7.4,
					7.6,
					7.8,
					8,
					8.2,
					8.4,
					8.6,
					8.8,
					9,
					9.2,
					9.4,
					9.6,
					9.8,
					10,
				},
				hide axis,
				ybar,
				bar width=5pt,
				ymin=0,
				every node near coord/.append style={rotate=90, anchor=west},
				width=\textwidth, 
				enlarge x limits={abs=0.5*\pgfplotbarwidth},
				height=9cm, 
				width=16cm,
				axis x line*=bottom, axis y line*=left
				]
				\addplot [fill=red] coordinates {
					(0,0)
					(0.2,0)
					(0.4,0)
					(0.6,0)
					(0.8,0)
					(1,0)
					(1.2,0)
					(1.4,0)
					(1.6,0)
					(1.8,0)
					(2,0)
					(2.2,0)
					(2.4,0)
					(2.6,0)
					(2.8,0)
					(3,0)
					(3.2,0)
					(3.4,0)
					(3.6,0)
					(3.8,0)
					(4,0)
					(4.2,0)
					(4.4,0)
					(4.6,0)
					(4.8,0)
					(5,0)
					(5.2,0)
					(5.4,0)
					(5.6,0)
					(5.8,0)
					(6,0)
					(6.2,0)
					(6.4,0)
					(6.6,0)
					(6.8,0)
					(7,0)
					(7.2,0)
					(7.4,0)
					(7.6,0)
					(7.8,0)
					(8,0)
					(8.2,0)
					(8.4,0)
					(8.6,0)
					(8.8,0)
					(9,0)
					(9.2,0)
					(9.4,55000)
					(9.6,0)
					(9.8,0)
					(10,0)
				};	
			\end{axis}
			
			\begin{axis}[ 
				symbolic x coords={
					0,
					0.2,
					0.4,
					0.6,
					0.8,
					1,
					1.2,
					1.4,
					1.6,
					1.8,
					2,
					2.2,
					2.4,
					2.6,
					2.8,
					3,
					3.2,
					3.4,
					3.6,
					3.8,
					4,
					4.2,
					4.4,
					4.6,
					4.8,
					5,
					5.2,
					5.4,
					5.6,
					5.8,
					6,
					6.2,
					6.4,
					6.6,
					6.8,
					7,
					7.2,
					7.4,
					7.6,
					7.8,
					8,
					8.2,
					8.4,
					8.6,
					8.8,
					9,
					9.2,
					9.4,
					9.6,
					9.8,
					10,
				},
				hide axis,
				ybar,
				bar width=5pt,
				ymin=0,
				every node near coord/.append style={rotate=90, anchor=west},
				width=\textwidth, 
				enlarge x limits={abs=0.5*\pgfplotbarwidth},
				height=9cm, 
				width=16cm,
				axis x line*=bottom, axis y line*=left
				]
				\addplot [fill=black] coordinates {
					(0,0)
					(0.2,0)
					(0.4,0)
					(0.6,0)
					(0.8,0)
					(1,0)
					(1.2,0)
					(1.4,0)
					(1.6,0)
					(1.8,0)
					(2,0)
					(2.2,0)
					(2.4,0)
					(2.6,0)
					(2.8,0)
					(3,0)
					(3.2,0)
					(3.4,0)
					(3.6,0)
					(3.8,0)
					(4,0)
					(4.2,0)
					(4.4,0)
					(4.6,0)
					(4.8,0)
					(5,0)
					(5.2,0)
					(5.4,0)
					(5.6,0)
					(5.8,0)
					(6,0)
					(6.2,0)
					(6.4,0)
					(6.6,0)
					(6.8,0)
					(7,0)
					(7.2,0)
					(7.4,0)
					(7.6,0)
					(7.8,0)
					(8,0)
					(8.2,0)
					(8.4,0)
					(8.6,55000)
					(8.8,0)
					(9,0)
					(9.2,0)
					(9.4,0)
					(9.6,0)
					(9.8,0)
					(10,0)
				};	
			\end{axis}
			
			\begin{axis}[
				ylabel={Number of Student},
				symbolic x coords={
					0,
					0.2,
					0.4,
					0.6,
					0.8,
					1,
					1.2,
					1.4,
					1.6,
					1.8,
					2,
					2.2,
					2.4,
					2.6,
					2.8,
					3,
					3.2,
					3.4,
					3.6,
					3.8,
					4,
					4.2,
					4.4,
					4.6,
					4.8,
					5,
					5.2,
					5.4,
					5.6,
					5.8,
					6,
					6.2,
					6.4,
					6.6,
					6.8,
					7,
					7.2,
					7.4,
					7.6,
					7.8,
					8,
					8.2,
					8.4,
					8.6,
					8.8,
					9,
					9.2,
					9.4,
					9.6,
					9.8,
					10,
				},
				xtick=data,
				x tick label style={rotate=90,anchor=east},
				ybar,
				bar width=5pt,
				ymin=0,
				nodes near coords,   
				every node near coord/.append style={rotate=90, anchor=west},
				width=\textwidth, 
				enlarge x limits={abs=0.5*\pgfplotbarwidth},
				height=9cm, 
				width=16cm,
				axis x line*=bottom, axis y line*=left
				]
				\addplot [fill=blue] coordinates {
					(0,6)
					(0.2,1)
					(0.4,0)
					(0.6,12)
					(0.8,38)
					(1,135)
					(1.2,362)
					(1.4,1022)
					(1.6,1966)
					(1.8,3410)
					(2,5852)
					(2.2,8734)
					(2.4,12421)
					(2.6,16601)
					(2.8,20744)
					(3,24243)
					(3.2,27391)
					(3.4,30085)
					(3.6,32408)
					(3.8,33932)
					(4,34393)
					(4.2,35450)
					(4.4,34992)
					(4.6,34848)
					(4.8,33738)
					(5,32844)
					(5.2,31110)
					(5.4,29487)
					(5.6,27915)
					(5.8,26531)
					(6,24751)
					(6.2,23612)
					(6.4,22103)
					(6.6,21201)
					(6.8,20276)
					(7,19147)
					(7.2,18972)
					(7.4,18126)
					(7.6,17946)
					(7.8,17575)
					(8,17236)
					(8.2,17512)
					(8.4,17022)
					(8.6,17120)
					(8.8,16526)
					(9,15248)
					(9.2,13262)
					(9.4,9620)
					(9.6,5460)
					(9.8,2222)
					(10,494)			
				};	
				
			\end{axis}
		\end{tikzpicture}
	\end{center}
	\caption{English score spectrum of Vietnamese students in 2023.}
	\label{fig:English_2023}
\end{figure*}

Figures~\ref{fig:English_2019}-\ref{fig:English_2023} show the English score spectrum of Vietnamese students in 2019-2023.

\begin{figure*}[h!]
	\begin{center}
		\begin{tikzpicture}
			\begin{axis}[
				legend style={at={(0.5,-0.125)}, 	
					anchor=north,legend columns=-1}, 
				symbolic x coords={
					0,
					0.2,
					0.4,
					0.6,
					0.8,
					1,
					1.2,
					1.4,
					1.6,
					1.8,
					2,
					2.2,
					2.4,
					2.6,
					2.8,
					3,
					3.2,
					3.4,
					3.6,
					3.8,
					4,
					4.2,
					4.4,
					4.6,
					4.8,
					5,
					5.2,
					5.4,
					5.6,
					5.8,
					6,
					6.2,
					6.4,
					6.6,
					6.8,
					7,
					7.2,
					7.4,
					7.6,
					7.8,
					8,
					8.2,
					8.4,
					8.6,
					8.8,
					9,
					9.2,
					9.4,
					9.6,
					9.8,
					10,
				},
				hide axis,
				ybar,
				bar width=5pt,
				ymin=0,
				ymax=10,
				every node near coord/.append style={rotate=90, anchor=west},
				width=\textwidth, 
				enlarge x limits={abs=0.5*\pgfplotbarwidth},
				height=9cm, 
				width=16cm,
				axis x line*=bottom, axis y line*=left
				]
				\addplot [fill=green] coordinates {
					(0,0)
				};
				\addplot [fill=red] coordinates {
					(5,0)
				};
				\addplot [fill=black] coordinates {
					(8,0)
				};	
				\addplot [fill=blue] coordinates {
					(10,0)
				};	
				\legend{ChatGPT, BingChat, Bard, Vietnamese students }	
			\end{axis}
			
			\begin{axis}[
				symbolic x coords={
					0,
					0.2,
					0.4,
					0.6,
					0.8,
					1,
					1.2,
					1.4,
					1.6,
					1.8,
					2,
					2.2,
					2.4,
					2.6,
					2.8,
					3,
					3.2,
					3.4,
					3.6,
					3.8,
					4,
					4.2,
					4.4,
					4.6,
					4.8,
					5,
					5.2,
					5.4,
					5.6,
					5.8,
					6,
					6.2,
					6.4,
					6.6,
					6.8,
					7,
					7.2,
					7.4,
					7.6,
					7.8,
					8,
					8.2,
					8.4,
					8.6,
					8.8,
					9,
					9.2,
					9.4,
					9.6,
					9.8,
					10,
				},
				hide axis,
				x tick label style={rotate=90,anchor=east},
				ybar,
				bar width=5pt,
				ymin=0,
				every node near coord/.append style={rotate=90, anchor=west},
				width=\textwidth, 
				enlarge x limits={abs=0.5*\pgfplotbarwidth},
				height=9cm, 
				width=16cm,
				axis x line*=bottom, axis y line*=left
				]
				\addplot [fill=green] coordinates {
					(0,0)
					(0.2,0)
					(0.4,0)
					(0.6,0)
					(0.8,0)
					(1,0)
					(1.2,0)
					(1.4,0)
					(1.6,0)
					(1.8,0)
					(2,0)
					(2.2,0)
					(2.4,0)
					(2.6,0)
					(2.8,0)
					(3,0)
					(3.2,0)
					(3.4,0)
					(3.6,0)
					(3.8,0)
					(4,0)
					(4.2,0)
					(4.4,0)
					(4.6,0)
					(4.8,0)
					(5,0)
					(5.2,0)
					(5.4,0)
					(5.6,0)
					(5.8,0)
					(6,0)
					(6.2,0)
					(6.4,0)
					(6.6,0)
					(6.8,0)
					(7,0)
					(7.2,0)
					(7.4,0)
					(7.6,55000)
					(7.8,0)
					(8,0)
					(8.2,0)
					(8.4,0)
					(8.6,0)
					(8.8,0)
					(9,0)
					(9.2,0)
					(9.4,0)
					(9.6,0)
					(9.8,0)
					(10,0)
					
				};	
			\end{axis}
			
			\begin{axis}[ 
				symbolic x coords={
					0,
					0.2,
					0.4,
					0.6,
					0.8,
					1,
					1.2,
					1.4,
					1.6,
					1.8,
					2,
					2.2,
					2.4,
					2.6,
					2.8,
					3,
					3.2,
					3.4,
					3.6,
					3.8,
					4,
					4.2,
					4.4,
					4.6,
					4.8,
					5,
					5.2,
					5.4,
					5.6,
					5.8,
					6,
					6.2,
					6.4,
					6.6,
					6.8,
					7,
					7.2,
					7.4,
					7.6,
					7.8,
					8,
					8.2,
					8.4,
					8.6,
					8.8,
					9,
					9.2,
					9.4,
					9.6,
					9.8,
					10,
				},
				hide axis,
				ybar,
				bar width=5pt,
				ymin=0,
				every node near coord/.append style={rotate=90, anchor=west},
				width=\textwidth, 
				enlarge x limits={abs=0.5*\pgfplotbarwidth},
				height=9cm, 
				width=16cm,
				axis x line*=bottom, axis y line*=left
				]
				\addplot [fill=red] coordinates {
					(0,0)
					(0.2,0)
					(0.4,0)
					(0.6,0)
					(0.8,0)
					(1,0)
					(1.2,0)
					(1.4,0)
					(1.6,0)
					(1.8,0)
					(2,0)
					(2.2,0)
					(2.4,0)
					(2.6,0)
					(2.8,0)
					(3,0)
					(3.2,0)
					(3.4,0)
					(3.6,0)
					(3.8,0)
					(4,0)
					(4.2,0)
					(4.4,0)
					(4.6,0)
					(4.8,0)
					(5,0)
					(5.2,0)
					(5.4,0)
					(5.6,0)
					(5.8,0)
					(6,0)
					(6.2,0)
					(6.4,0)
					(6.6,0)
					(6.8,0)
					(7,0)
					(7.2,0)
					(7.4,0)
					(7.6,0)
					(7.8,0)
					(8,0)
					(8.2,0)
					(8.4,0)
					(8.6,0)
					(8.8,0)
					(9,0)
					(9.2,55000)
					(9.4,0)
					(9.6,0)
					(9.8,0)
					(10,0)
				};	
			\end{axis}
			
			\begin{axis}[ 
				symbolic x coords={
					0,
					0.2,
					0.4,
					0.6,
					0.8,
					1,
					1.2,
					1.4,
					1.6,
					1.8,
					2,
					2.2,
					2.4,
					2.6,
					2.8,
					3,
					3.2,
					3.4,
					3.6,
					3.8,
					4,
					4.2,
					4.4,
					4.6,
					4.8,
					5,
					5.2,
					5.4,
					5.6,
					5.8,
					6,
					6.2,
					6.4,
					6.6,
					6.8,
					7,
					7.2,
					7.4,
					7.6,
					7.8,
					8,
					8.2,
					8.4,
					8.6,
					8.8,
					9,
					9.2,
					9.4,
					9.6,
					9.8,
					10,
				},
				hide axis,
				ybar,
				bar width=5pt,
				ymin=0,
				every node near coord/.append style={rotate=90, anchor=west},
				width=\textwidth, 
				enlarge x limits={abs=0.5*\pgfplotbarwidth},
				height=9cm, 
				width=16cm,
				axis x line*=bottom, axis y line*=left
				]
				\addplot [fill=black] coordinates {
					(0,0)
					(0.2,0)
					(0.4,0)
					(0.6,0)
					(0.8,0)
					(1,0)
					(1.2,0)
					(1.4,0)
					(1.6,0)
					(1.8,0)
					(2,0)
					(2.2,0)
					(2.4,0)
					(2.6,0)
					(2.8,0)
					(3,0)
					(3.2,0)
					(3.4,0)
					(3.6,0)
					(3.8,0)
					(4,0)
					(4.2,0)
					(4.4,0)
					(4.6,0)
					(4.8,0)
					(5,0)
					(5.2,0)
					(5.4,0)
					(5.6,0)
					(5.8,0)
					(6,0)
					(6.2,0)
					(6.4,0)
					(6.6,0)
					(6.8,0)
					(7,0)
					(7.2,0)
					(7.4,0)
					(7.6,0)
					(7.8,0)
					(8,0)
					(8.2,55000)
					(8.4,0)
					(8.6,0)
					(8.8,0)
					(9,0)
					(9.2,0)
					(9.4,0)
					(9.6,0)
					(9.8,0)
					(10,0)
				};	
			\end{axis}
		
			\begin{axis}[
				ylabel={Number of Student},
				symbolic x coords={
					0,
					0.2,
					0.4,
					0.6,
					0.8,
					1,
					1.2,
					1.4,
					1.6,
					1.8,
					2,
					2.2,
					2.4,
					2.6,
					2.8,
					3,
					3.2,
					3.4,
					3.6,
					3.8,
					4,
					4.2,
					4.4,
					4.6,
					4.8,
					5,
					5.2,
					5.4,
					5.6,
					5.8,
					6,
					6.2,
					6.4,
					6.6,
					6.8,
					7,
					7.2,
					7.4,
					7.6,
					7.8,
					8,
					8.2,
					8.4,
					8.6,
					8.8,
					9,
					9.2,
					9.4,
					9.6,
					9.8,
					10,
				},
				xtick=data,
				x tick label style={rotate=90,anchor=east},
				ybar,
				bar width=5pt,
				ymin=0,
				nodes near coords,   
				every node near coord/.append style={rotate=90, anchor=west},
				width=\textwidth, 
				enlarge x limits={abs=0.5*\pgfplotbarwidth},
				height=9cm, 
				width=16cm,
				axis x line*=bottom, axis y line*=left
				]
				\addplot [fill=blue] coordinates {
					(0,0)
					(0.2,0)
					(0.4,5)
					(0.6,33)
					(0.8,123)
					(1,469)
					(1.2,1324)
					(1.4,2864)
					(1.6,5952)
					(1.8,10310)
					(2,16722)
					(2.2,23685)
					(2.4,31481)
					(2.6,37599)
					(2.8,42348)
					(3,45297)
					(3.2,45755)
					(3.4,44476)
					(3.6,41861)
					(3.8,39542)
					(4,36385)
					(4.2,33410)
					(4.4,30588)
					(4.6,27458)
					(4.8,24979)
					(5,22630)
					(5.2,20989)
					(5.4,18710)
					(5.6,17283)
					(5.8,15464)
					(6,14288)
					(6.2,13145)
					(6.4,12173)
					(6.6,11343)
					(6.8,10405)
					(7,9834)
					(7.2,9274)
					(7.4,8552)
					(7.6,7990)
					(7.8,7612)
					(8,7108)
					(8.2,6970)
					(8.4,6416)
					(8.6,6045)
					(8.8,5378)
					(9,4845)
					(9.2,3968)
					(9.4,3133)
					(9.6,1976)
					(9.8,939)
					(10,299)
					
				};	
				
			\end{axis}
		\end{tikzpicture}
	\end{center}
	\caption{English score spectrum of Vietnamese students in 2019.}
	\label{fig:English_2019}
\end{figure*}
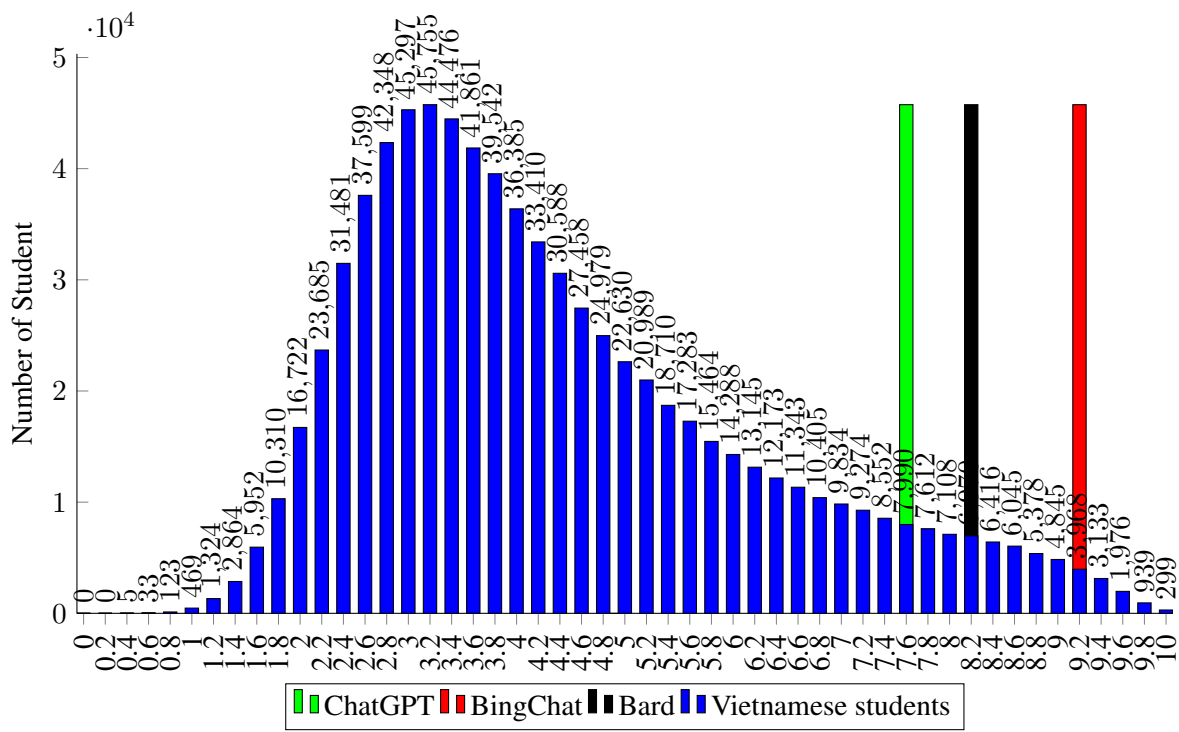

\begin{figure*}[h!]
	\begin{center}
		\begin{tikzpicture}
			\begin{axis}[
				legend style={at={(0.5,-0.125)}, 	
					anchor=north,legend columns=-1}, 
				symbolic x coords={
					0,
					0.2,
					0.4,
					0.6,
					0.8,
					1,
					1.2,
					1.4,
					1.6,
					1.8,
					2,
					2.2,
					2.4,
					2.6,
					2.8,
					3,
					3.2,
					3.4,
					3.6,
					3.8,
					4,
					4.2,
					4.4,
					4.6,
					4.8,
					5,
					5.2,
					5.4,
					5.6,
					5.8,
					6,
					6.2,
					6.4,
					6.6,
					6.8,
					7,
					7.2,
					7.4,
					7.6,
					7.8,
					8,
					8.2,
					8.4,
					8.6,
					8.8,
					9,
					9.2,
					9.4,
					9.6,
					9.8,
					10,
				},
				hide axis,
				ybar,
				bar width=5pt,
				ymin=0,
				ymax=10,
				every node near coord/.append style={rotate=90, anchor=west},
				width=\textwidth, 
				enlarge x limits={abs=0.5*\pgfplotbarwidth},
				height=9cm, 
				width=16cm,
				axis x line*=bottom, axis y line*=left
				]
				\addplot [fill=green] coordinates {
					(0,0)
				};
				\addplot [fill=red] coordinates {
					(5,0)
				};
				\addplot [fill=black] coordinates {
					(7,0)
				};	
				\addplot [fill=blue] coordinates {
					(10,0)
				};	
				\legend{ChatGPT, BingChat,Vietnamese students }	
			\end{axis}
			
			\begin{axis}[
				symbolic x coords={
					0,
					0.2,
					0.4,
					0.6,
					0.8,
					1,
					1.2,
					1.4,
					1.6,
					1.8,
					2,
					2.2,
					2.4,
					2.6,
					2.8,
					3,
					3.2,
					3.4,
					3.6,
					3.8,
					4,
					4.2,
					4.4,
					4.6,
					4.8,
					5,
					5.2,
					5.4,
					5.6,
					5.8,
					6,
					6.2,
					6.4,
					6.6,
					6.8,
					7,
					7.2,
					7.4,
					7.6,
					7.8,
					8,
					8.2,
					8.4,
					8.6,
					8.8,
					9,
					9.2,
					9.4,
					9.6,
					9.8,
					10,
				},
				hide axis,
				x tick label style={rotate=90,anchor=east},
				ybar,
				bar width=5pt,
				ymin=0,
				every node near coord/.append style={rotate=90, anchor=west},
				width=\textwidth, 
				enlarge x limits={abs=0.5*\pgfplotbarwidth},
				height=9cm, 
				width=16cm,
				axis x line*=bottom, axis y line*=left
				]
				\addplot [fill=green] coordinates {
					(0,0)
					(0.2,0)
					(0.4,0)
					(0.6,0)
					(0.8,0)
					(1,0)
					(1.2,0)
					(1.4,0)
					(1.6,0)
					(1.8,0)
					(2,0)
					(2.2,0)
					(2.4,0)
					(2.6,0)
					(2.8,0)
					(3,0)
					(3.2,0)
					(3.4,0)
					(3.6,0)
					(3.8,0)
					(4,0)
					(4.2,0)
					(4.4,0)
					(4.6,0)
					(4.8,0)
					(5,0)
					(5.2,0)
					(5.4,0)
					(5.6,0)
					(5.8,0)
					(6,0)
					(6.2,0)
					(6.4,0)
					(6.6,0)
					(6.8,0)
					(7,0)
					(7.2,0)
					(7.4,0)
					(7.6,0)
					(7.8,0)
					(8,0)
					(8.2,0)
					(8.4,0)
					(8.6,55000)
					(8.8,0)
					(9,0)
					(9.2,0)
					(9.4,0)
					(9.6,0)
					(9.8,0)
					(10,0)
					
				};	
			\end{axis}
			
			\begin{axis}[ 
				symbolic x coords={
					0,
					0.2,
					0.4,
					0.6,
					0.8,
					1,
					1.2,
					1.4,
					1.6,
					1.8,
					2,
					2.2,
					2.4,
					2.6,
					2.8,
					3,
					3.2,
					3.4,
					3.6,
					3.8,
					4,
					4.2,
					4.4,
					4.6,
					4.8,
					5,
					5.2,
					5.4,
					5.6,
					5.8,
					6,
					6.2,
					6.4,
					6.6,
					6.8,
					7,
					7.2,
					7.4,
					7.6,
					7.8,
					8,
					8.2,
					8.4,
					8.6,
					8.8,
					9,
					9.2,
					9.4,
					9.6,
					9.8,
					10,
				},
				hide axis,
				ybar,
				bar width=5pt,
				ymin=0,
				every node near coord/.append style={rotate=90, anchor=west},
				width=\textwidth, 
				enlarge x limits={abs=0.5*\pgfplotbarwidth},
				height=9cm, 
				width=16cm,
				axis x line*=bottom, axis y line*=left
				]
				\addplot [fill=red] coordinates {
					(0,0)
					(0.2,0)
					(0.4,0)
					(0.6,0)
					(0.8,0)
					(1,0)
					(1.2,0)
					(1.4,0)
					(1.6,0)
					(1.8,0)
					(2,0)
					(2.2,0)
					(2.4,0)
					(2.6,0)
					(2.8,0)
					(3,0)
					(3.2,0)
					(3.4,0)
					(3.6,0)
					(3.8,0)
					(4,0)
					(4.2,0)
					(4.4,0)
					(4.6,0)
					(4.8,0)
					(5,0)
					(5.2,0)
					(5.4,0)
					(5.6,0)
					(5.8,0)
					(6,0)
					(6.2,0)
					(6.4,0)
					(6.6,0)
					(6.8,0)
					(7,0)
					(7.2,0)
					(7.4,0)
					(7.6,0)
					(7.8,0)
					(8,0)
					(8.2,0)
					(8.4,0)
					(8.6,0)
					(8.8,0)
					(9,0)
					(9.2,0)
					(9.4,0)
					(9.6,55000)
					(9.8,0)
					(10,0)
				};	
			\end{axis}
			\begin{axis}[ 
				symbolic x coords={
					0,
					0.2,
					0.4,
					0.6,
					0.8,
					1,
					1.2,
					1.4,
					1.6,
					1.8,
					2,
					2.2,
					2.4,
					2.6,
					2.8,
					3,
					3.2,
					3.4,
					3.6,
					3.8,
					4,
					4.2,
					4.4,
					4.6,
					4.8,
					5,
					5.2,
					5.4,
					5.6,
					5.8,
					6,
					6.2,
					6.4,
					6.6,
					6.8,
					7,
					7.2,
					7.4,
					7.6,
					7.8,
					8,
					8.2,
					8.4,
					8.6,
					8.8,
					9,
					9.2,
					9.4,
					9.6,
					9.8,
					10,
				},
				hide axis,
				ybar,
				bar width=5pt,
				ymin=0,
				every node near coord/.append style={rotate=90, anchor=west},
				width=\textwidth, 
				enlarge x limits={abs=0.5*\pgfplotbarwidth},
				height=9cm, 
				width=16cm,
				axis x line*=bottom, axis y line*=left
				]
				\addplot [fill=black] coordinates {
					(0,0)
					(0.2,0)
					(0.4,0)
					(0.6,0)
					(0.8,0)
					(1,0)
					(1.2,0)
					(1.4,0)
					(1.6,0)
					(1.8,0)
					(2,0)
					(2.2,0)
					(2.4,0)
					(2.6,0)
					(2.8,0)
					(3,0)
					(3.2,0)
					(3.4,0)
					(3.6,0)
					(3.8,0)
					(4,0)
					(4.2,0)
					(4.4,0)
					(4.6,0)
					(4.8,0)
					(5,0)
					(5.2,0)
					(5.4,0)
					(5.6,0)
					(5.8,0)
					(6,0)
					(6.2,0)
					(6.4,0)
					(6.6,0)
					(6.8,0)
					(7,0)
					(7.2,0)
					(7.4,0)
					(7.6,0)
					(7.8,0)
					(8,0)
					(8.2,0)
					(8.4,0)
					(8.6,0)
					(8.8,0)
					(9,0)
					(9.2,0)
					(9.4,55000)
					(9.6,0)
					(9.8,0)
					(10,0)
				};	
			\end{axis}
		
			\begin{axis}[
				ylabel={Number of Student},
				symbolic x coords={
					0,
					0.2,
					0.4,
					0.6,
					0.8,
					1,
					1.2,
					1.4,
					1.6,
					1.8,
					2,
					2.2,
					2.4,
					2.6,
					2.8,
					3,
					3.2,
					3.4,
					3.6,
					3.8,
					4,
					4.2,
					4.4,
					4.6,
					4.8,
					5,
					5.2,
					5.4,
					5.6,
					5.8,
					6,
					6.2,
					6.4,
					6.6,
					6.8,
					7,
					7.2,
					7.4,
					7.6,
					7.8,
					8,
					8.2,
					8.4,
					8.6,
					8.8,
					9,
					9.2,
					9.4,
					9.6,
					9.8,
					10,
				},
				xtick=data,
				x tick label style={rotate=90,anchor=east},
				ybar,
				bar width=5pt,
				ymin=0,
				nodes near coords,   
				every node near coord/.append style={rotate=90, anchor=west},
				width=\textwidth, 
				enlarge x limits={abs=0.5*\pgfplotbarwidth},
				height=9cm, 
				width=16cm,
				axis x line*=bottom, axis y line*=left
				]
				\addplot [fill=blue] coordinates {
					(0,13)
					(0.2,1)
					(0.4,5)
					(0.6,30)
					(0.8,102)
					(1,392)
					(1.2,1099)
					(1.4,2263)
					(1.6,4431)
					(1.8,8283)
					(2,12427)
					(2.2,18461)
					(2.4,24565)
					(2.6,29001)
					(2.8,33167)
					(3,35670)
					(3.2,37285)
					(3.4,37335)
					(3.6,36730)
					(3.8,35597)
					(4,34627)
					(4.2,32682)
					(4.4,31295)
					(4.6,29713)
					(4.8,27816)
					(5,26106)
					(5.2,23932)
					(5.4,22050)
					(5.6,19904)
					(5.8,18173)
					(6,16453)
					(6.2,14850)
					(6.4,13674)
					(6.6,12482)
					(6.8,11427)
					(7,10650)
					(7.2,10106)
					(7.4,9461)
					(7.6,9403)
					(7.8,8779)
					(8,8543)
					(8.2,8062)
					(8.4,7478)
					(8.6,6653)
					(8.8,5735)
					(9,4532)
					(9.2,3325)
					(9.4,2253)
					(9.6,1367)
					(9.8,672)
					(10,225)
					
				};	
				
			\end{axis}
		\end{tikzpicture}
	\end{center}
	\caption{English score spectrum of Vietnamese students in 2020.}
	\label{fig:English_2020}
\end{figure*}
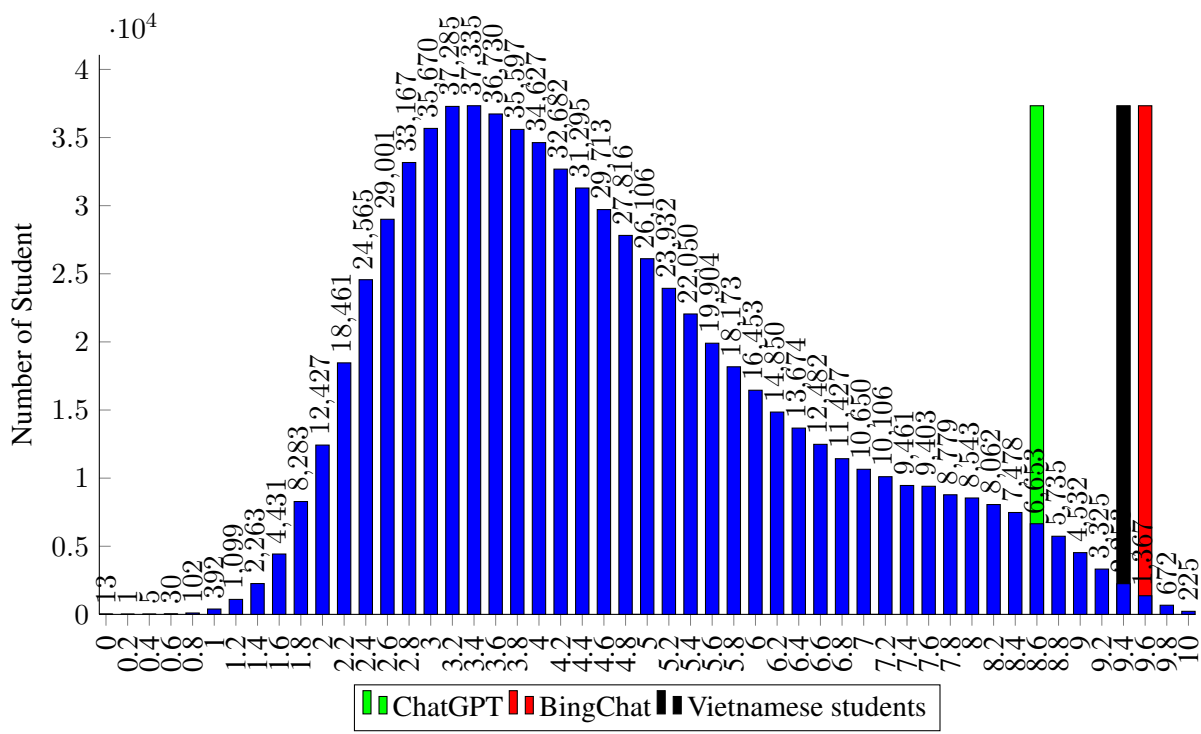

\end{document}